\documentclass[preprint,12pt]{elsarticle}

\date{}

\usepackage{caption}
\usepackage{subcaption}
\usepackage{multirow}
\usepackage{amsmath,amssymb,bm}
\usepackage{algorithm}
\usepackage{algorithmic}
\usepackage{xcolor}
\usepackage{float}
\usepackage[a4paper,margin=2.5cm]{geometry}
\usepackage{tikz}
\usetikzlibrary{positioning, arrows.meta}
\usepackage{booktabs}
\usepackage{graphicx}
\usepackage{caption}
\usetikzlibrary{positioning, shapes, arrows.meta, fit, backgrounds, calc}

\usepackage{amssymb}
\usepackage{amsmath}
\usepackage{tikz}
\usepackage{rotating}
\usepackage{hyperref}
\usepackage{amsmath, amssymb}

\usepackage{tikz}
\usetikzlibrary{arrows.meta, positioning, shapes, decorations.pathreplacing, calc}
\usepackage{comment}

\usepackage{booktabs}
\usepackage{siunitx}
\usepackage[percent]{overpic}
\usepackage[utf8]{inputenc}
\usepackage{siunitx}
\usepackage{graphicx}

\usepackage{subcaption}

\definecolor{myGreen}{rgb}{0.4501, 0.6784,0.2745}

\definecolor{myblue}{rgb}{0.705882, 0.78039, 0.90588}

\definecolor{mywhite}{rgb}{1,1,1}
\definecolor{blue2}{rgb}{0.7098,0.78039,0.913725}

\definecolor{blueshape}{rgb}{0.670589,0.74902,0.901961}

\definecolor{greenframe}{rgb}{0.77255,1,0.77255}

\definecolor{greenabstract}{rgb}{0.55691,0.851,0.451}

\definecolor{blueabstract}{rgb}{0.27451,0.6941,0.88235}

\journal{International Journal of Heat and Mass Transfer}

\begin{document}

\begin{frontmatter}





\title{HeatGen: A Guided Diffusion Framework for Multiphysics Heat Sink Design Optimization}

\author[label1]{Hadi Keramati\corref{cor1}}
\ead{hadi.keramati@ubc.ca}

\author[TUD]{Morteza Sadeghi}
\ead{morteza.sadeghi@nmf.tu-darmstadt.de}

\author[label1]{Rajeev K. Jaiman}
\ead{rjaiman@mech.ubc.ca}
\cortext[cor1]{Corresponding author: hadi.keramati@ubc.ca}

\affiliation[label1]{organization={Department of Mechanical Engineering},
            addressline={University of British Columbia}, 
            city={Vancouver}, 
            state={BC},
            country={Canada}}
\address[TUD]{Institute for Nano- and Microfluidics,
Department of Mechanical Engineering, Technical University of Darmstadt, Darmstadt, Germany.}
\begin{abstract}

This study presents a generative optimization framework based on a guided denoising diffusion probabilistic model (DDPM) that leverages surrogate gradients to generate heat sink designs minimizing pressure drop while maintaining surface temperatures below a specified threshold.
Geometries are represented using boundary representations of multiple fins, and a multi-fidelity approach is employed to generate training data.
Using this dataset, along with vectors representing the boundary representation geometries, we train a denoising diffusion probabilistic model to generate heat sinks with characteristics consistent with those observed in the data.
We train two different residual neural networks to predict the pressure drop and surface temperature for each geometry. We use the gradients of these surrogate models with respect to the design variables to guide the geometry generation process toward satisfying the low-pressure and surface temperature constraints.
This inference-time guidance directs the generative process toward heat sink designs that not only prevent overheating but also achieve lower pressure drops compared to traditional optimization methods such as CMA-ES.
In contrast to traditional black-box optimization approaches, our method is scalable, provided sufficient training data is available. 
Unlike traditional topology optimization methods, once the model is trained and the heat sink world model is saved, inference under new constraints (e.g., temperature) is computationally inexpensive and does not require retraining.
Samples generated using the guided diffusion model achieve pressure drops up to 10 percent lower than the limits obtained by traditional black-box optimization methods. This work represents a step toward building a foundational generative model for electronics cooling.

\end{abstract}

\end{frontmatter}



\section{Introduction}
\label{sec1}
Heat transfer devices are fundamental to a wide range of energy-system applications, from large-scale HVAC systems \cite{gilmore2021manifold} to compact electronics cooling \cite{hetsroni2002uniform} and medical device thermal management \cite{naphon2019ann}. Their design involves navigating a high-dimensional space governed by the nonlinear Navier--Stokes and convection--diffusion equations, where fluid flow, conduction, and convection interact in complex ways \cite{hachem2024reinforcement}. Optimal design must balance competing objectives which is mainly minimizing maximum temperature while reducing pressure drop or pumping power \cite{mekki2021genetic, sahin2024deep}.

A classical approach to navigating the design space for multiple objectives, borrowed from structural and mechanical design, is the use of adjoint-based solvers, which have been applied to heat sink and heat exchanger optimization~\cite{sun20203d}. This method computes the sensitivities of the objective with respect to geometric representations and uses them to guide gradient-based updates~\cite{jameson2003aerodynamic, gallorini2023multi}. While effective for smooth, well-behaved problems, adjoint methods often struggle with the nonlinear and multiphysics nature of thermo-fluid systems~\cite{keramati2022deep}. Local minima and poor gradient conditioning can trap the optimization process. Moreover, formulating a gradient operator for the design space is both challenging and time-consuming, as it necessitates constructing an adjoint operator (or system) specifically tailored to the forward model and objective function to enable efficient gradient computation~\cite{plessix2006review}.

Another major family of approaches is black-box optimization, including evolutionary algorithms such as genetic algorithms \cite{golberg1989genetic}, Covariance matrix adaptation evolution strategy \cite{hansen2001completely}, and particle swarm optimization \cite{kennedy1995particle}. These methods, typically coupled with computational fluid dynamics (CFD) solvers or surrogate-CFD hybrids, allow faster iteration and exploration of more complex geometry representations \cite{arie2017air,keramati2023deep}. However, their primary limitation lies in scalability and sample efficiency, since increasing geometric complexity and design freedom (e.g., fin shape, count, or overall layout) make the computational cost prohibitive. Consequently, many studies limit geometric freedom, for instance, by constraining fin spacing or using two-dimensional simplifications. Different geometric representations, including voxel- and pixel-based formulations, allow high design freedom, but face similar computational barriers due to strongly coupled fluid–thermal physics. 

%
In recent years, machine learning (ML) and artificial intelligence (AI) have emerged as promising tools to optimize the design of heat transfer devices \cite{jahanbakhsh2026physics,song2025design}. Reinforcement-learning-based and boundary-representation frameworks have been used to automate fin-layout and channel-geometry design, optimizing for increased heat transfer and reduced pressure drop \cite{keramati2022generative}. Surrogate models, often trained using ML, are also widely trained to accelerate performance evaluations \cite{togun2025machine}. Black-box optimizers can leverage these surrogates even when they are non-differentiable, whereas adjoint methods require differentiability or analytical sensitivities~\cite{krishnamoorthy2023diffusion, koo2014shape}. Neural operators have also been used to predict optimized heat-transfer topologies, but primarily for conduction problems and within density-based formulations~\cite{yuan2025method}.

Building on these surrogate-driven approaches, recent advances in generative modeling have aimed to automate the geometric synthesis process in a data-driven approach, in which an offline dataset is used to train the generative model for various engineering applications~\cite{chen2022inverse, chen2023gan,wang2022ih}. Generative Adversarial Networks (GANs) have been explored for conduction-based heat sink design, revealing the potential for data-driven shape generation~\cite{drake2024quantize}. GANs suffer from training instability and limited representational capacity, making them difficult to extend to convection-dominated or fully coupled fluid–thermal systems \cite{krishnamoorthy2023diffusion}. Therefore, a more stable and physically consistent generative framework is required to capture the nonlinear dependencies between geometry, flow, and heat transfer.

To overcome the limitations of GANs for design optimization frameworks, diffusion models have emerged as a robust and mathematically principled alternative to generative learning \cite{kumar2020model}. These models construct a gradual noise-injection process in which data are progressively perturbed by Gaussian noise and then reconstructed through a learned reverse diffusion process \cite{ho2020denoising}. 
The discrete formulation, known as the Denoising Diffusion Probabilistic Model (DDPM) \cite{ho2020denoising}, and its continuous counterpart based on stochastic differential equations \cite{song2020score,song2021maximum}, provide stable training dynamics and high sample diversity without adversarial instability. Owing to their probabilistic nature, diffusion models can be conditioned on external gradients or surrogate objectives, enabling a targeted generation toward high-performance or physically feasible designs \cite{dhariwal2021diffusion,ho2022classifier}. This controllable sampling capability makes diffusion frameworks particularly appealing for multi-objective optimization problems, including complex thermo-fluidic systems such as heat sinks. Guidance from surrogate models has been successfully applied in other engineering domains, such as aerospace and marine design \cite{bagazinski2023shipgen, keramati2025reward}. However, despite the potential benefits of generative models for thermo-fluidic design, no prior study has explored their application in this context.

Motivated by these capabilities, this work introduces HeatGen, a diffusion-based generative design framework for multiphysics heat sink optimization. The framework integrates a denoising diffusion probabilistic model with differentiable surrogate models for pressure drop and surface temperature, guiding the generation process toward thermally efficient and hydraulically optimized geometries which are represented using Bézier-curve boundary parameterization. The main backbone of these models are reconstruction models, such as U-Net, that learn to reconstruct a matrix or a vector representation. In this work, we employ a one-dimensional U-Net backbone known as Seq-U-Net, which has proven to be efficient for sequence modeling tasks \cite{stoller2019seq}. This architecture is well-suited for modeling structured geometric representations of Bézier fin vectors in heat sink design.

The proposed framework formulates heat-sink optimization as a generative sampling problem subject to two competing physical objectives:
(i) constraining the maximum temperature below a specified threshold to prevent overheating, and 
(ii) minimizing the pressure drop to enhance flow efficiency. 
A multi-fidelity data-generation pipeline is developed in three-dimensional space, where heat-sink geometries are parameterized using Bézier-based fin representations. A denoising diffusion probabilistic model is first pre-trained to learn the distribution of feasible designs, after which gradient-based sampling, guided by two surrogate neural networks trained on the initial dataset, biases the generation toward low-pressure-drop regions of the design space that also satisfy the overheating-prevention constraint. 

 To the best of our knowledge, this study represents the first application of probabilistic diffusion models to the optimization of heat-sink geometries under fully coupled multiphysics and multi-objective conditions. The proposed framework introduces a scalable, data-driven pathway for the heat transfer community to perform surrogate-informed generative design under convection–conduction coupling, where conventional adjoint or gradient-based solvers often become numerically unstable. By embedding differentiable surrogate models directly into the diffusion inference loop, HeatGen transforms traditional design-of-experiments and reduced-order modeling concepts into a unified framework for physics-aware generative design automation.

The remainder of this paper is organized as follows: Section 2 presents the theoretical background of diffusion-based generative modeling. Section 3 describes the multi-fidelity data-generation and thermal–fluid simulation methodology. Section 4 outlines the geometric representation and data flow through the network architectures that construct the diffusion model. Section 5 introduces the gradient-guided optimization strategy. Section 6 discusses the results and validation, and Section 7 concludes with key findings and perspectives for future research.

\section{Background and Theoretical Framework}
\label{sec:background}
Before describing the data-driven implementation, we briefly summarize the mathematical foundation of diffusion models for the proposed framework. Diffusion-based generative models have emerged as a powerful probabilistic framework for learning and sampling from complex, high-dimensional data distributions through a gradual stochastic transformation between structured data and isotropic noise \cite{sohl2015deep}.  Unlike adversarial or autoencoding models, which implicitly or explicitly map between data and latent variables via deterministic or adversarial training, diffusion models construct an explicit Markov chain that incrementally destroys the structure in the data and then learns to reverse this process \cite{sohl2015deep, ho2020denoising}.  
The resulting framework provides a principled formulation rooted in nonequilibrium thermodynamics, where data generation corresponds to a learned reverse-time diffusion process that reconstructs structured samples from pure noise.

\subsection{Generative Diffusion Framework}
\label{subsec:framework}

Let \(x_0 \in \mathbb{R}^d\) denote a clean data sample drawn from an unknown data distribution \(p_{\mathrm{data}}(x_0)\).  
A denoising diffusion probabilistic model defines a sequence of latent variables \(\{x_t\}_{t=1}^T\) of the same dimensionality, connected through a fixed forward diffusion process \(q(x_{1:T} \mid x_0)\).  
This process iteratively corrupts the data by adding Gaussian noise with a small variance \(\beta_t\) at each discrete timestep \(t\in\{1,\ldots,T\}\):
\begin{equation}
q(x_t \mid x_{t-1}) = \mathcal{N}\!\left(x_t;\, \sqrt{1 - \beta_t}\,x_{t-1},\, \beta_t I\right),
\label{eq:forward}
\end{equation}
where \(I\) is the identity matrix.  
The noise schedule \(\{\beta_t\}_{t=1}^T\) is typically monotonically increasing to ensure a smooth transition from data to noise.  

By recursively applying \eqref{eq:forward}, the marginal distribution of \(x_t\) conditioned on \(x_0\) can be written in closed form using the cumulative product of \(\alpha_t = 1 - \beta_t\):
\begin{equation}
x_t = \sqrt{\bar{\alpha}_t}\,x_0 + \sqrt{1 - \bar{\alpha}_t}\,\epsilon, 
\quad \text{where } \bar{\alpha}_t = \prod_{s=1}^{t}\alpha_s,\;
\epsilon \sim \mathcal{N}(0, I).
\label{eq:marginal}
\end{equation}
This equation implies that any intermediate noisy state \(x_t\) can be sampled directly from \(x_0\) without explicitly simulating the entire chain up to \(t\).  
As \(t \to T\), \(\bar{\alpha}_t \to 0\), and the data distribution asymptotically approaches a standard normal prior \(q(x_T) \approx \mathcal{N}(0, I)\), providing a simple, tractable starting point for the reverse process.
The forward diffusion thus defines a smooth trajectory that gradually pushes the data manifold toward an isotropic Gaussian, enabling a reversible probabilistic mapping between structured and unstructured representations.

\subsection{Reverse Denoising Dynamics}
\label{subsec:dynamics}

The generative process seeks to invert the forward diffusion by learning a reverse-time Markov chain
\begin{equation}
p_\theta(x_{0:T}) = p(x_T)\prod_{t=1}^T p_\theta(x_{t-1}\mid x_t),
\end{equation}
where the transition kernels \(p_\theta(x_{t-1}\mid x_t)\) are parameterized by a neural network with parameters \(\theta\).  
Each reverse step is modeled as a Gaussian distribution:
\begin{equation}
p_\theta(x_{t-1} \mid x_t)
= \mathcal{N}\!\left(x_{t-1};\, \mu_\theta(x_t, t),\, \sigma_t^2 I\right),
\label{eq:reverse}
\end{equation}
where the variance \(\sigma_t^2\) may be fixed (e.g., \(\beta_t\)) or learned \cite{nichol2021improved}. In this study, we use a fixed variance.
To ensure consistency with the forward process, the mean \(\mu_\theta(x_t, t)\) is reparameterized in terms of the model’s predicted noise \(\epsilon_\theta(x_t, t)\):
\begin{equation}
\mu_\theta(x_t, t)
= \frac{1}{\sqrt{\alpha_t}}
\!\left(
x_t - \frac{\beta_t}{\sqrt{1 - \bar{\alpha}_t}}\,
\epsilon_\theta(x_t, t)
\right).
\label{eq:mu}
\end{equation}
This form enforces that the model learns to estimate the added noise \(\epsilon\) at each step, which is computationally more stable and statistically efficient than directly predicting \(x_0\) or \(x_{t-1}\).  
A single reverse update is then given by
\begin{equation}
x_{t-1} = \mu_\theta(x_t, t) + \sigma_t z,
\quad z \sim \mathcal{N}(0, I),
\label{eq:reconstruction}
\end{equation}
and repeating this procedure for \(t = T, T-1, \ldots, 1\) gradually transforms pure Gaussian noise \(x_T \sim \mathcal{N}(0, I)\) into a realistic sample \(x_0 \sim p_\theta(x_0)\).  
The iterative nature of this process provides a flexible mechanism for conditional or guided generation, since additional gradients or conditioning signals (e.g., rewards, classifiers, or physical constraints) can be introduced at each timestep.

\subsection{Learning Objective}
\label{subsec:training}

Model parameters \(\theta\) are optimized via maximum likelihood estimation to approximate the true data distribution \(p_{\mathrm{data}}(x_0)\).  
However, directly maximizing the marginal likelihood \(\log p_\theta(x_0)\) is intractable, so training instead minimizes a variational upper bound on the negative log-likelihood, i.e., the variational lower bound (VLB), as given by:
\begin{equation}
\mathcal{L}_{\mathrm{VLB}}
=
\mathbb{E}_q\!\left[
-\log\frac{p_\theta(x_{0:T})}{q(x_{1:T}\mid x_0)}
\right]
=
\sum_{t=1}^T \mathbb{E}_q\!\big[
D_{\mathrm{KL}}\!\big(q(x_{t-1}\!\mid x_t, x_0)\,\|\,p_\theta(x_{t-1}\!\mid x_t)\big)
\big]
- \log p_\theta(x_0\!\mid x_1),
\label{eq:vlb}
\end{equation}
where \(D_{\mathrm{KL}}\) denotes the Kullback--Leibler divergence.  

As shown in \cite{ho2020denoising}, with a fixed variance schedule, this objective simplifies to a noise prediction loss, yielding the practical training criterion:
\begin{equation}
\mathcal{L}_{\text{DDPM}}
=
\mathbb{E}_{x_0,\,\epsilon,\,t}
\!\left[
\|\epsilon - \epsilon_\theta(x_t, t)\|_2^2
\right],
\quad x_t = \sqrt{\bar{\alpha}_t}x_0 + \sqrt{1-\bar{\alpha}_t}\,\epsilon.
\label{eq:ddpm_loss}
\end{equation}
This formulation teaches the model to estimate the Gaussian noise component at each timestep, effectively denoising intermediate latent variables.  
Unlike the common use of diffusion models trained purely on image pixels, the 1D Seq-U-Net used in this study operates in a structured parametric space of Bézier geometry vectors. This enables the model to learn spatial correlations along fin contours, directly linking latent variables to physically meaningful geometric parameters.

\section{Dataset Generation Methodology}
\label{sec:datasetandtp}
\begin{figure}[h!]
    \centering
    \includegraphics[width=1.0\textwidth]{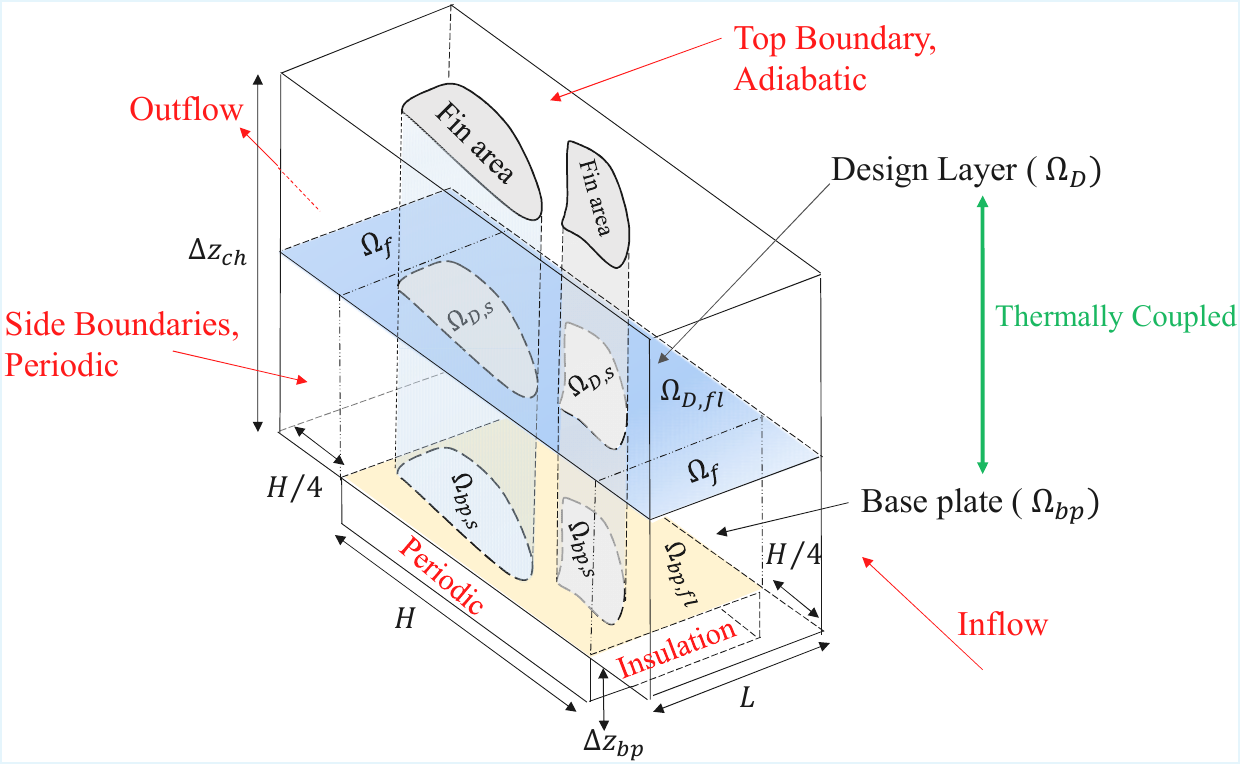}
    \caption{ A 3D schematic view of the problem along with the simplified pseudo-3D model. It represents the conversion from the full 3D model to two 2D models, with boundary conditions indicated.}
    \label{fig:schematic}
\end{figure}

In order to train diffusion and surrogate models, a large dataset is required for training, validation, and testing. A total of 27,500 samples were generated using a pseudo-3D numerical model to simulate the coupled thermo-fluidic behavior of the heat sink. The adopted computational domain, schematized in Fig. \ref{fig:schematic}, simplifies the full 3D problem into two thermally coupled 2D domains: a solid base plate and a thermo-fluidic design layer. Further explanations of how the solid fin regions, \(\Omega_{d,s}\), shown in Fig.~\ref{fig:schematic}, are constructed using composite Bézier curves are provided in \ref{app:bezier}~\cite{sadeghi2025multi}.

 The geometric parameters for the 2D model and the corresponding 3D validation model are based on the characteristic length $H=10 \text{ mm}$. This defines the base plate thickness as $\Delta{z}_{bp}=H/8$, the channel height as $\Delta{z}_{ch}=1.5H$, and the width of the modeled section as $L=0.5H$.

\subsection{Fluid Dynamics Model}

The fluid is considered to be incompressible and the flow as two-dimensional and  steady in the $xy$-plane. The governing equations consist of the continuity and steady-state Navier–Stokes equations:
\begin{align}
\frac{\partial u}{\partial x} + \frac{\partial v}{\partial y} &= 0, \\[6pt]
\rho_f \left( u \frac{\partial u}{\partial x} + v \frac{\partial u}{\partial y} \right)
&= - \frac{\partial p}{\partial x}
+ \mu_f \left( 
\frac{\partial^2 u}{\partial x^2}
+ \frac{\partial^2 u}{\partial y^2}
\right), \\[6pt]
\rho_f \left( u \frac{\partial v}{\partial x} + v \frac{\partial v}{\partial y} \right)
&= - \frac{\partial p}{\partial y}
+ \mu_f \left(
\frac{\partial^2 v}{\partial x^2}
+ \frac{\partial^2 v}{\partial y^2}
\right).
\end{align}
where $\rho_f$ is the fluid density, $u$ and $v$ are the fluid velocity in the x and y directions, respectively, $p$ is the pressure field, and $\mu_f$ is the dynamic viscosity of the fluid.

\subsection{Thermal Model and Pseudo-3D Coupling}
The thermal model is constructed using two thermally interacting two-dimensional domains, namely the solid base plate and the thermo-fluidic design layer whose coupling yields an equivalent three-dimensional thermal representation. In the solid base plate ($\Omega_{bp}$), the 2D heat conduction equation is solved for the temperature field $T_{bp}$, assuming uniform volumetric heat generation ($\dot{Q}_{prod}$) and accounting for the heat transfer to the upper layer as a sink term:

\begin{equation}
- k_s \left(
\frac{\partial^2 T_{bp}}{\partial x^2}
+
\frac{\partial^2 T_{bp}}{\partial y^2}
\right)
=
\frac{\dot{Q}_{prod}}{V_{bp}}
-
\frac{\dot{q}_{\text{inter}}}{\Delta z_{bp}}
\label{eq:solid_base_plate}
\end{equation}
here, $k_s$ is the solid thermal conductivity, $\dot{Q}_{prod}$ is the total prescribed heat rate, $V_{bp}$ is the base plate volume, and $\Delta z_{bp}$ is the base plate height.

In the design layer, the temperature field $T$ is solved. This layer is subdivided into the external fluid region ($\Omega_f$), the solid fin regions ($\Omega_{d,s}$), and the internal fluid regions ($\Omega_{d,fl}$).
In the external thermofluid layer ($\Omega_f$), the 2D thermal convection-diffusion equation given as follows is solved:
\begin{equation}
\rho_f c_f \left( 
u \frac{\partial T}{\partial x} 
+ v \frac{\partial T}{\partial y} 
\right)
- k_f \left(
\frac{\partial^2 T}{\partial x^2}
+  \frac{\partial^2 T}{\partial y^2}
\right)
= 0
\end{equation}
In the design domain, the governing equations for the solid fin parts ($\Omega_{d,s}$) and fluid parts ($\Omega_{d,fl}$) are written as:
\begin{equation}
- k_s \left(
\frac{\partial^2 T}{\partial x^2}
+
\frac{\partial^2 T}{\partial y^2}
\right)
=
\frac{\dot{q}_{\text{inter,s}}}{\Delta z_{ch}}
\quad \text{in } \Omega_{d,s}
\end{equation}
\begin{equation}
\rho_f c_f \left(
u \frac{\partial T}{\partial x}
+
v \frac{\partial T}{\partial y}
\right)
-
k_f \left(
\frac{\partial^2 T}{\partial x^2}
+
\frac{\partial^2 T}{\partial y^2}
\right)
=
\frac{\dot{q}_{\text{inter,f}}}{\Delta z_{ch}}
\quad \text{in } \Omega_{d,fl}
\end{equation}
where $c_f$ is the specific fluid heat capacity and $k_f$ is the fluid thermal conductivity. The source and sink terms $\dot{q}_{\text{inter}}$ handle the thermal coupling between the base plate ($T_{bp}$) and the design layer ($T$). These terms model the vertical (z-direction) heat transfer.
\begin{align*}
\dot{q}_{\text{inter,s}} &= h_s (T_{bp}-T) && \text{(solid-to-solid conduction)} \\
\dot{q}_{\text{inter,f}} &= h_f (T_{bp}-T) && \text{(solid-to-fluid convection)}
\end{align*}
here, $h_f$ is the convective heat transfer coefficient and $h_s$ represents the vertical conductive heat transfer within the fins, assumed to take an averaged value based on the reference data. The sink term $\dot{q}_{\text{inter}}$ within the base plate equation, Equation ~\ref{eq:solid_base_plate}, is spatially defined and set equal to $\dot{q}_{\text{inter,s}}$ within the fin-attached areas ($\Omega_{bp,s}$) and equal to $\dot{q}_{\text{inter,f}}$ within the fluid-contact areas ($\Omega_{bp,fl}$).

The governing equations are subjected to the following boundary conditions:
A uniform inlet velocity of $u_{in}=1 \text{ m/s}$ is prescribed for the fluid flow at the inlet, while the outflow boundary is set with a zero-pressure condition $p=0$. The repeating nature of the heat sink configuration is represented by imposing periodic boundary conditions on the lateral boundaries. For the heat transfer problem, the fluid enters at a fixed temperature $T_{in}=298.15 \text{ K}$ and a convective outflow condition $n \cdot \nabla T = 0$ at the outlet. Thermal periodic boundary conditions are imposed in a similar fashion on the side boundaries. The front and back surfaces of the solid base plate are taken to be adiabatic.

To evaluate the performance of the heat sink, the total pressure drop and the average surface temperature are calculated. The pressure drop is defined as the difference between the area-averaged inlet and outlet pressures:
\begin{equation}
\Delta \overline{p} =
\overline{p}_{inlet}
-
\overline{p}_{outlet}
\label{eq:pressure_difference_new}
\end{equation}
The average value of pressure or temperature over a boundary $\Gamma$ can be calculated with the general formula: \begin{equation}
\overline{\Psi}_{\Gamma} =
\frac{\displaystyle\int_{\Gamma} \Psi\,d\Xi}{\displaystyle\int_{\Gamma} d\Xi}
\label{eq:avg_new}
\end{equation}
where $\Psi$ represents the field variable ($p$ or $T$) and $d\Xi$ is the differential line element or area element ($dA$), depending on the dimension of $\Gamma$.  For instance, the average surface temperature is: 
\begin{equation}
\overline{T}_{\Gamma} =
\frac{\displaystyle\int_{\Gamma} T\,d\Xi}{\displaystyle\int_{\Gamma} d\Xi}.
\label{eq:Tavg_new}
\end{equation}

In order to evaluate the generate data, we employ an automated multi-platform workflow integrating Python, MATLAB, and COMSOL. The parameter sets are first generated in the Python environment and then passed to MATLAB. Through the COMSOL–MATLAB programming interface, the corresponding \texttt{.mph} model is constructed and simulated within COMSOL. The simulation results are then returned to MATLAB and subsequently transferred back to Python for data generation.

\section{Design Vector Representation and 
Reconstruction}
Each design sample corresponds to two Bézier fin geometries represented by a
48-dimensional vector:
\begin{equation}
x_0 =
[x_{11}, x_{12}, \ldots, x_{1,12},\;
 y_{11}, y_{12}, \ldots, y_{1,12},\;
 x_{21}, x_{22}, \ldots, x_{2,12},\;
 y_{21}, y_{22}, \ldots, y_{2,12}]
\in \mathbb{R}^{48}.
\end{equation}

\begin{figure}[htbp]
  \centering
  \begin{tikzpicture}[scale=0.67, transform shape,
      node distance=0.7cm,
      every node/.style={font=\footnotesize, align=center},
      block/.style={rectangle, draw, thick, minimum width=1.6cm, minimum height=0.6cm},
      smallblock/.style={rectangle, draw, thick, minimum width=0.27cm, minimum height=0.5cm},
      arrow/.style={-Latex, thick},
      line/.style={-Latex, thick, draw=black, shorten >=2pt, shorten <=2pt}
    ]

  \node[block, fill=pink!20, minimum height=2.5cm, minimum width=2.8cm] (latent) {};

  \begin{scope}[shift={(latent.south west)}, xshift=1cm, yshift=1.25cm]
    \node[smallblock, fill=orange!20, minimum width=0.6cm, minimum height=0.35cm] (lres1) {\rotatebox{90}{MidResblock1}};
    \node[smallblock, fill=yellow!20, minimum width=0.6cm, minimum height=0.35cm, right=0.1cm of lres1] (lres2) {\rotatebox{90}{MidResblock2}};
  \end{scope}
  \node[font=\footnotesize] (latent_left_text) at ($(latent)+(-3.4cm,0)$) {Latent space $(B, 256, 6)$};
  \node[font=\footnotesize] (latent_right_text) at ($(latent)+(2.4cm,0)$) {$(B, 256, 6)$};
  \node[smallblock, fill=blue!20]   (res11) at ($(latent)+(2.5cm,2.5cm)$) {\rotatebox{90}{conv 1D(3,stride1)}};
  \node[smallblock, fill=orange!20] (res12) at ($(res11)+(0.6cm,0)$) {\rotatebox{90}{Resblock1}};
  \node[smallblock, fill=yellow!20] (res13) at ($(res12)+(0.6cm,0)$) {\rotatebox{90}{Resblock2}};
  \node[draw, line, inner sep=0.2cm, fit=(res11)(res12)(res13)] (g1) {};
  \node[fill=white] (de_block1) at ($(latent)+(6.5cm,2.5cm)$) {Upsample 1\\(B, 128, 12)};

  \node[smallblock, fill=blue!20] (res21) at ($(latent)+(-2.5cm,2.5cm)$) {\rotatebox{90}{conv 1D(4,stride2)}};
  \node[smallblock, fill=orange!20] (res22) at ($(res21)+(-0.6cm,0)$) {\rotatebox{90}{Resblock1}};
  \node[smallblock, fill=yellow!20] (res23) at ($(res22)+(-0.6cm,0)$) {\rotatebox{90}{Resblock2}};
  \node[fill=white] (en_block2) at ($(latent)+(-6.5cm,2.5cm)$) {Downsample 3\\(B, 128, 12)};
  \node[draw, line, inner sep=0.2cm, fit=(res21)(res22)(res23)] (g2) {};

  \node[smallblock, fill=blue!20] (res31) at ($(res13)+(1.5cm,2.5cm)$) {\rotatebox{90}{conv 1D(1,stride1)}};
  \node[smallblock, fill=orange!20] (res32) at ($(res31)+(0.6cm,0)$) {\rotatebox{90}{Resblock1}};
  \node[smallblock, fill=yellow!20] (res33) at ($(res32)+(0.6cm,0)$) {\rotatebox{90}{Resblock2}};
  \node[fill=white] (up_down1) at ($(res33)+(3cm,0)$) {Upsample 2\\(B, 64, 24)};
  \node[draw, line, inner sep=0.2cm, fit=(res31)(res32)(res33)] (g3) {};

  \node[smallblock, fill=blue!20] (res41) at ($(res23)+(-1.5cm,2.5cm)$) {\rotatebox{90}{conv 1D(4,stride2)}};
  \node[smallblock, fill=orange!20] (res42) at ($(res41)+(-0.6cm,0)$) {\rotatebox{90}{Resblock1}};
  \node[smallblock, fill=yellow!20] (res43) at ($(res42)+(-0.6cm,0)$) {\rotatebox{90}{Resblock2}};
  \node[fill=white] (up_down2) at ($(res43)+(-3cm,0)$) {Downsample2\\(B, 64, 24)};
  \node[draw, line, inner sep=0.2cm, fit=(res41)(res42)(res43)] (g4) {};

  \node[smallblock, fill=blue!20] (res51) at ($(res33)+(1.5cm,2.5cm)$) {\rotatebox{90}{conv 1D(3,stride1)}};
  \node[smallblock, fill=orange!20] (res52) at ($(res51)+(0.6cm,0)$) {\rotatebox{90}{Resblock1}};
  \node[smallblock, fill=yellow!20] (res53) at ($(res52)+(0.6cm,0)$) {\rotatebox{90}{Resblock2}};
  \node[fill=white] (de_block2) at ($(res53)+(3cm,0)$) {Upsample 3\\(B, 64, 48)};
  \node[draw, line, inner sep=0.2cm, fit=(res51)(res52)(res53)] (g5) {};

  \node[smallblock, fill=blue!20] (res61) at ($(res43)+(-1.5cm,2.5cm)$) {\rotatebox{90}{conv 1D(4,stride2)}};
  \node[smallblock, fill=orange!20] (res62) at ($(res61)+(-0.6cm,0)$) {\rotatebox{90}{Resblock1}};
  \node[smallblock, fill=yellow!20] (res63) at ($(res62)+(-0.6cm,0)$) {\rotatebox{90}{Resblock2}};
  \node[fill=white] (enc_block1) at ($(res63)+(-3cm,0)$) {Downsample 1\\(B, 64, 48)};
  \node[draw, line, inner sep=0.2cm, fit=(res61)(res62)(res63)] (g6) {};

  \node[smallblock, fill=orange!20] (res71) at ($(res53)+(1.5cm,2.5cm)$) {\rotatebox{90}{Group norm}};
  \node[smallblock, fill=yellow!20] (res72) at ($(res71)+(0.6cm,0)$) {\rotatebox{90}{SiLU}};
  \node[smallblock, fill=blue!20] (res73) at ($(res72)+(0.5cm,0)$) {\rotatebox{90}{conv 1D(3,stride 1)}};
  \node[fill=white] (eps_block) at ($(res73)+(1.5cm,0)$) {$\epsilon_t$\\(B, 1, 48)};
  \node[draw, line, inner sep=0.2cm, fit=(res71)(res72)(res73)] (g7) {};

  \node[smallblock, fill=blue!20] (res81) at ($(res63)+(-1.5cm,2.5cm)$) {\rotatebox{90}{conv 1D(3,stride 1)}};
  \node[fill=white] (xblock) at ($(res81)+(-1.5cm,0)$) {$x_t$\\(B, 1, 48)};
  \node[draw, line, inner sep=0.1cm, fit=(res81)] (g8) {};

  \node[circle, draw, fill=white, minimum size=0.1cm] (plus1) at ($(latent)+(0,2.5cm)$) {+};
  \node[circle, draw, fill=white, minimum size=0.1cm] (plus2) at ($(plus1)+(0,2.5cm)$) {+};
  \node[circle, draw, fill=white, minimum size=0.1cm] (plus3) at ($(plus2)+(0,2.5cm)$) {+};

  \draw[arrow] (g8.south) -- (g6.west);
  \draw[arrow] (g6.south) -- (g4.west);
  \draw[arrow] (g4.south) -- (g2.west);
  \draw[arrow] (g2.south) -- (latent);
  \draw[arrow] (latent) -- (g1.south);
  \draw[arrow] (g1.east) -- (g3.south);

  \draw[arrow] (g3.east) -- (g5.south);
  \draw[arrow] (g5.east) -- (g7.south);

  \draw[arrow] (g2.east) -- (g1.west);
  \draw[arrow] (g4.east) -- (g3.west);
  \draw[arrow] (g6.east) -- (g5.west);

  \end{tikzpicture}
  \caption{Architecture of the 1D U-Net for geometry reconstruction of fin arrays}
  \label{fig:1dUnet}
\end{figure}
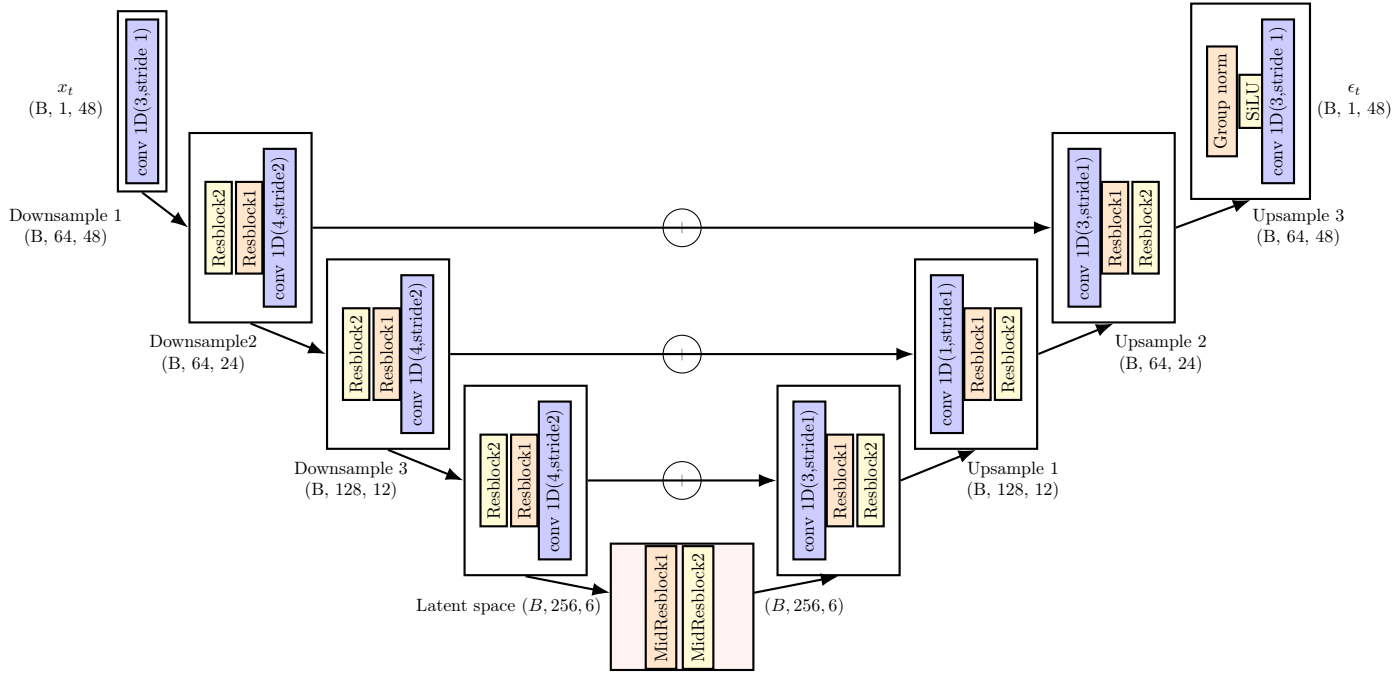

\noindent
Each design is defined by the coordinates of
\(\text{Fin 1: } (x_{1,i}, y_{1,i})_{i=1}^{12}\) and 
\(\text{Fin 2: } (x_{2,i}, y_{2,i})_{i=1}^{12}\), 
where \((x_{k,i}, y_{k,i})\) denotes the coordinates of the point \(i^{\text{th}}\) point of the \(k^{\text{th}}\) curve. During preprocessing, all coordinates are extracted in canonical order,
and each feature is standardized according to $\tilde{x}_i = \frac{x_i - \mu_i}{\sigma_i},$ yielding standardized vectors
$\tilde{x}_0 \in \mathbb{R}^{48}$.
Each standardized design vector is interpreted as a one-dimensional sequence of
length~48:
\begin{equation}
\tilde{x}_0 = [\tilde{x}_1, \tilde{x}_2, \ldots, \tilde{x}_{48}]^\top.
\end{equation}
This representation allows the model to interpret the Bézier geometry as a one-dimensional structured sequence, where local correlations between consecutive points capture the smoothness and geometric consistency of the fin profile within the neural network framework. This representation is used within a one-dimensional U-Net architecture, as shown in Fig.~\ref{fig:1dUnet}, to reconstruct multiple fin geometries as a sequential model embedded in a denoising diffusion probabilistic models.

\begin{figure}[H]
    \centering

    \tikzstyle{block2} = [rectangle, rounded corners, minimum width=0.8cm, minimum height=1.3cm, 
                          text centered, draw=black, fill=white!20]
    \tikzstyle{blockgray} = [rectangle, rounded corners, minimum width=1cm, minimum height=1.5cm, 
                             text centered, draw=black, fill=black!10]
    \tikzstyle{arrow} = [thick,->,>=stealth]
    \tikzstyle{dashedarrow} = [thick, dashed,->,>=stealth]

    \resizebox{0.95\textwidth}{!}{
    \begin{tikzpicture}[node distance=2.8cm]
        \node (xT)   [block2] {$x_T$};
        \node (xTm1) [block2, right=2.8cm of xT]   {$x_{T-1}$};
        \node (xTm2) [block2, right=2.9cm of xTm1] {$x_{T-2}$};
        \node (x1)   [block2, right=3.6cm of xTm2] {$x_1$};
        \node (x0)   [blockgray, right=2.8cm of x1] {$x_0$};
        \node at ($(xTm2.east) + (0.5,-0.1)$) {$\cdots$};

        \draw [dashedarrow]($(xTm1.west) + (0,+0.2)$) -- ($(xT.east) + (0,+0.2)$);
        \draw [dashedarrow]($(xTm2.west) + (0,+0.2)$) -- node[above] {$q(x_{T-1}|x_{T-2})$}($(xTm1.east) + (0,+0.2)$);
        \draw [dashedarrow]($(x1.west) + (0,+0.2)$)   -- node[above] {$q(x_{2}|x_{1})$}($(xTm2.east) + (1,+0.17)$);
        \draw [dashedarrow]($(x0.west)+(0,+0.2)$)     -- node[above] {$q(x_{1}|x_{0})$}($(x1.east)+(0,+0.2)$);

        \draw [arrow] (xT)   -- node[below] {} (xTm1);
        \draw [arrow] (xTm1) -- node[below] {$p_{\theta}(x_{T-2} | x_{T-1})$} (xTm2);
        \draw [arrow] ($(xTm2.east) + (1,0)$) -- node[below] {$p_{\theta}(x_{1}|x_{2})$} (x1);
        \draw [arrow] (x1)   -- node[below] {$p_{\theta}(x_{0} | x_{1})$} (x0);
    \end{tikzpicture}
    }

    \caption{Graphical model of the denoising diffusion probabilistic model (DDPM). 
    Dashed arrows denote the forward diffusion process $q(x_t|x_{t-1})$, while solid arrows 
    represent the learned reverse denoising process $p_\theta(x_{t-1}|x_t)$.}
    \label{fig:actual_ddpm}
\end{figure}
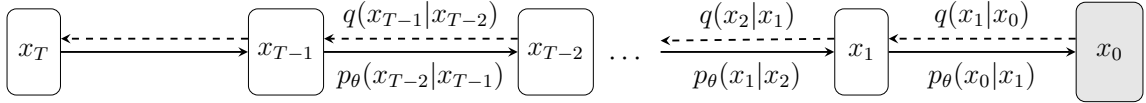

Each downsampling block is shown in Fig.~\ref{fig:downsample-encoder} which divides the sequence length by two and increases the channel size.
\begin{figure}[h]
\centering
\tikzset{
  block/.style={rectangle, draw, thick, minimum width=2.5cm, minimum height=0.8cm, align=center},
  smallblock/.style={rectangle, draw, thick, minimum width=2.5cm, minimum height=0.6cm, align=center},
  arrow/.style={-Latex, thick},
  brace/.style={decorate, decoration={brace, amplitude=5pt}},
  every node/.style={font=\rmfamily},
  line/.style={-Latex, thick, draw=black, shorten >=2pt, shorten <=2pt},
}

\begin{tikzpicture}[node distance=0.8cm]

\node (input) {Downsample};

\node[block, below=of input, fill=orange!20] (block1) {Resblock1};
\node[block, below=of block1, fill=yellow!20] (block2) {Resblock2};
\node[block, below=of block2, fill=blue!20] (conv) {conv 1D(4, stride2)};
\node[below=of conv, font=\small] (out) {Output};
\node[right=0.75cm of input, font=\small] (skip) {Skip};

\node[draw, line, inner sep=0.5cm, fit=(block1)(block2)(conv), label=above:{\small }] {};

\draw[arrow] (input) -- (block1);
\draw[arrow] (block1) -- (block2);
\draw[arrow] (block2) -- (conv);
\draw[arrow] (conv.south) -- ++(0,-0.8);
\draw[arrow] ($ (input) + (0cm,-0.5cm) $) -- ++(2.75,0);

\end{tikzpicture}
\caption{Illustration of the downsampling module consisting of two residual blocks followed by a convolution layer with stride 2.}
\label{fig:downsample-encoder}
\end{figure}
Each ResBlock shown in Fig.~\ref{fig:downsample-encoder} contains several operations, detailed in Fig.~\ref{fig:ResBlock}, where the timestep embedding from the reverse diffusion process is integrated to inform the U-Net of its current timestep.
\begin{figure}[h!]
    \centering
    \resizebox{0.9\textwidth}{!}{%
        \begin{tikzpicture}[
            node distance=0.5cm,
            font=\scriptsize,
            scale=0.9,
            every node/.style={transform shape}
        ]

        \tikzset{
            block/.style={rectangle, draw, thick, minimum width=1.8cm, minimum height=0.6cm, align=center, font=\footnotesize},
            smallblock/.style={rectangle, draw, thick, minimum width=1.8cm, minimum height=0.45cm, align=center, font=\footnotesize},
            arrow/.style={-Latex, thick},
            every node/.style={font=\rmfamily},
            line/.style={-Latex, thick, draw=black, shorten >=2pt, shorten <=2pt},
        }

        \node[block, fill=orange!20] (cov) {Conv1D(3)};
        \node[block, below= of cov, fill=yellow!20] (grpn) {Group norm};
        \node[block, below= of grpn, fill=pink!20] (silv2) {SiLU};
        \node[block, below=1cm of silv2, fill=orange!20] (cov2) {Conv1D(3)};
        \node[block, below= of cov2, fill=yellow!20] (grpn2) {Group norm};
        \node[block, below= of grpn2, fill=green!20] (drop) {SilU+drop};
        \node[circle, draw, fill=white, minimum size=0.3cm, below= of drop] (plus2) {+};

        \node at ($(silv2)+(-1.5cm,-0.75cm)$)[circle, draw, fill=white, minimum size=0.3cm] (plus_in) {+};

        \node[block, left=of plus_in,fill=blue!20] (linear){Linear};
        \node[block, left=of linear,fill=pink!20] (silv){SiLU};
        \node[block, left=of silv, fill=pink!60] (time) {time embed};

        \draw[arrow] (time) -- (silv);
        \draw[arrow] (silv) -- (linear);
        \draw[arrow] (linear) -- (plus_in);
        \draw[arrow] (plus_in) -- ($ (silv2) + (0cm, -0.75cm) $);
        \draw[arrow] (cov) -- (grpn);
        \draw[arrow] (grpn) -- (silv2);
        \draw[arrow] (silv2) -- (cov2);
        \draw[arrow] (cov2) -- (grpn2);
        \draw[arrow] (grpn2) -- (drop);
        \draw[arrow] (drop) -- (plus2);

        \node[above=0.7cm of cov] (input){\small Input x (B, C\textsubscript{in},L)};
        \draw[arrow] (input) -- (cov);

        \node[left=0.15cm of cov] (lp){\small Local pattern extraction};
        \node[left=0.15cm of grpn] (norm){\small Channel normalization};
        \node[left=0.15cm of silv2] (act){\small Activation};
        \node[above=0.15cm of time] (tt){\small (0,...,T-1)};
        \node[right=2.5cm of plus_in] (skip){\small Skip};
        \node[above=0.2cm of lp] (tx){\small };

        \draw[arrow, rounded corners=1pt]
        ($ (input) + (0cm,-0.4cm) $) -- ++(1.3,0) |- (plus2.east);

        \end{tikzpicture}%
    } 

    \caption{One Resblock with skip connections and time embedding}
    \label{fig:ResBlock}
\end{figure}
The discussion so far has focused on the architecture of the 1D U-Net to reconstruct fin geometries and arrays based on the design vector and the corresponding timestep along the reverse diffusion process. The 1D U-Net is then used to predict the noise component at each step, as illustrated in Fig. \ref{fig:actual_ddpm} and explained in section \ref{subsec:training}, to generate a new set of design vectors that closely resemble those in the training dataset. Further details about the DDPM hyperparameters are available in \ref{app:ddpm}.

\section{Heat Sink Design Optimization with Gradient Guidance }

The proposed framework reformulates the heat sink optimization problem as a sampling task using a DDPM method and addresses it within a constrained optimization framework. The first objective is to reduce the average surface temperature of the base plate and the second objective is to minimize the pressure loss of the driven fluid. In this study, they are handled as a combined minimization and constraint problem:

\begin{align}
\text{Minimize:} \quad & \Delta \overline{p} \label{eq:optimization}\\
\text{Subject to:} \quad &  \overline{T}_{bp} \leq T_{\mathrm{fixed}} \nonumber\\
\end{align}
where \(\Delta\overline{p}\) denotes the pressure drop across the heat sink, and \(T_{\mathrm{fixed}}\) represents the temperature constraint value obtained from~\eqref{eq:pressure_difference_new} and~\eqref{eq:Tavg_new}, respectively. We use regression models trained on the dataset generated according to the procedure described in section~\ref{sec:datasetandtp} to obtain \(\Delta\overline{p}\) and \(T_{\mathrm{fixed}}\), along with their gradients with respect to \(\tilde{x}_0\). The details of these methods are further explained in the next section.

\subsection{Regression guidance }
In the standard DDPM, the reverse process is defined by Eq. \eqref{eq:reverse}, and the training is guided by $\nabla_{x_t}\log p_\theta(x_t)$, where $\sigma_t$ is fixed and $\mu_\theta(x_t,t)$ is reparameterized in terms of the model’s predicted noise $\epsilon_\theta(x_t, t)$. Classifier-based guidance~\cite{dhariwal2021diffusion} was introduced to change the gradients to guide the generation to a specific class of images, and the gradient is modified as
\begin{equation}
\nabla_{x_t}\log p_\text{guided}(x_t|y)
= \nabla_{x_t}\log p_\theta(x_t)
+ \lambda\,\nabla_{x_t}\log p(y|x_t),
\label{eq:classifier-guidance}
\end{equation}
where $\lambda$ controls the strength of the guidance, and $\nabla_{x_t}\log p(y|x_t)$ is provided by a pre-trained classifier. 

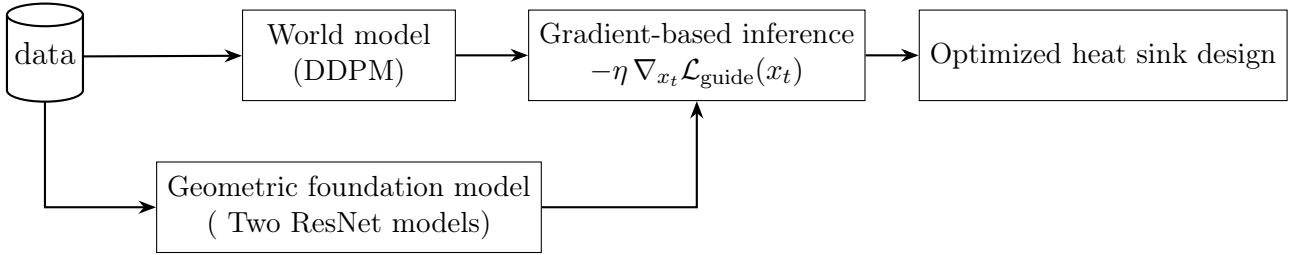
\begin{figure}[h]
    \centering
    \begin{tikzpicture}[
        node distance=2.8cm,
        >=Stealth,
        box/.style={draw, minimum width=2.8cm, minimum height=1.2cm, align=center, font=\small}
    ]

    \draw[thick, shift={(-5,-0.35)}] (0,2) ellipse (0.5 and 0.15);
    \draw[thick, shift={(-5,-0.35)}] (-0.5,2) -- (-0.5,1) 
        arc[start angle=180, end angle=360, x radius=0.5, y radius=0.15] -- (0.5,2);
    \node at (-5,1.15) {data};
    \node at (-1,-0.9) [box] (A) {Geometric foundation model\\( Two ResNet models)};
    \draw[->, thick] (-5,0.5) |- (A);

    \node[box,above=0.8cm of A] (B) {World model\\ (DDPM)};
    \node[box, right=.95cm of B] (C) {Gradient-based inference\\$- \eta\,\nabla_{x_t}\mathcal{L}_{\text{guide}}(x_t)$};

    \draw[->, thick] (-4.5,1.1) -- (B);
    \draw[->, thick] (A) -| (C);
    \draw[->, thick] (B) -- (C);

    \node[box, right=0.7cm of C] (D) {Optimized heat sink design};
    \draw[->, thick] (C) -- (D);

    \end{tikzpicture}
    \caption{Workflow of the model that generates optimized heat sink designs}
    \label{fig:gfmodel}
\end{figure}

In HeatGen, to guide the generative diffusion process toward physically meaningful designs, we integrate differentiable regression models into the denoising sampling loop. The objective is to synthesize geometry vectors $x \in \mathbb{R}^{48}$ that follow the constraints of Eq. \eqref{eq:optimization} to minimize the predicted pressure drop while maintaining the predicted temperature close to a prescribed target value $T_{\mathrm{fixed}}$. The process couples the learned generative prior from the DDPM with gradient-based guidance obtained from two pretrained ResNets that act as surrogates for pressure and temperature prediction. According to Eq. \eqref{eq:reverse}, \(p_\theta(x_t \mid x_{t+1})\) denotes the reverse diffusion model parameterized by \(\theta\) at timestep \(t\). We initialize the latent vector as \(x_T \sim \mathcal{N}(0, I)\) and progressively denoise it through \(T\) steps to obtain the generated geometry \(x_0\). For each sampling run, \(n_{\mathrm{samples}}\) geometries are generated in parallel $(x_t \in \mathbb{R}^{n_{\mathrm{samples}} \times 1 \times 48})$. Each sample represents a 48-dimensional shape parameterization vector defining two fin geometries. The diffusion process iterates backward through timesteps \(t = T-1, \dots, 0\). Instead of a discrete classifier, we use differentiable surrogate regression models for the continuous objective values of pressure drop $p_{\mathrm{pred}}(x_t)$ and base plate temperature $T_{\mathrm{pred}}(x_t)$. We define a guidance loss function:
\begin{equation}
\mathcal{L}_{\text{guided}}(x_t)
= \lambda_P\,p_{pred}(x_t)
+ \lambda_T\,\big(|T_{pred}(x_t)-T_{\text{fixed}}|-\Delta T\big)_+^2,
\label{eq:guide-loss}
\end{equation}
where $(\cdot)_+=\max(\cdot,0)$, and $\lambda_P$, $\lambda_T$ are scalar weights, and $\Delta T$ is the tolerance for the temperature. The gradient of this loss replaces the classifier gradient in Eq.~\eqref{eq:classifier-guidance} to form the modified gradient as follows:
\begin{equation}
\nabla_{x_t}\log p_\text{guided}(x_t)
= \nabla_{x_t}\log p_\theta(x_t)
- \eta\,\nabla_{x_t}\mathcal{L}_{\text{guide}}(x_t).
\label{eq:guided-score}
\end{equation}
Here $\eta>0$ is a scaling parameter and gradient with respect to the design vector is as follows:

\begin{equation}
\nabla_{x_{\mathrm{t}}} \mathcal{L}_{\text{guided}}
= 
\frac{\partial \mathcal{L}_{\text{guided}}}{\partial x_{\mathrm{t}}}
=
\lambda_P \frac{\partial \mathbb{E}[p_{\mathrm{pred}}]}{\partial x_{\mathrm{t}}}
+
\lambda_T \frac{\partial \mathbb{E}[\big(|T_{pred}(x_t)-T_{\text{fixed}}|-\Delta T\big)_+^2]}{\partial x_{\mathrm{t}}}.
\label{eq:gradient}
\end{equation}
In essence, the gradient information is used to adjust the current latent vector before each denoising step. This modification biases the generative trajectory toward geometries predicted to have low pressure and a temperature close to $T_{\mathrm{fixed}}$. Fig.~\ref{fig:gfmodel} shows the workflow and illustrates how data and regressors are used to construct the geometric foundation model. The same dataset is used to train a heat sink world model using the DDPM. Then, using gradients from the geometric foundation model to guide the inference of the world model, optimized heat sink designs are generated. The next section elaborates on how the two gradients for pressure and temperature are obtained using regression models.

\subsection{Regression models}
Two Residual Networks (ResNets) regressors \cite{he2016deep}, $p_{\mathrm{pred}} = f_P(\hat{x})$ and $T_{\mathrm{pred}} = f_T(\hat{x})$, serve as differentiable surrogates for the pressure drop and average surface temperature, respectively. The model architectures of the surrogate regressors are explained in detail in \ref{app:modelap}. Given the non-normalized geometry vector $x$, the inputs are quantile-normalized, denoted by $\hat{x}$. Once trained, we then compute the gradients of the loss from each surrogate model with respect to \(x\). PyTorch's automatic differentiation engine applies the chain rule through both ResNet surrogates, capturing their sensitivities with respect to the change in geometry vector. The differentiability of the surrogate models enables a physics-informed corrective direction in the latent space. Since gradients are computed with respect to the input geometry rather than model parameters, the surrogates remain frozen and computationally inexpensive. The resulting guided diffusion process thus generates designs aligned with specified performance targets. It is important to note that not all surrogate models can be utilized in this context, as the surrogate must be differentiable with respect to the design vector to provide guidance to the denoising process. This approach is therefore not applicable to non-differentiable surrogates or models with prohibitively high differentiation costs (e.g., graph neural networks).

\section{Results and Discussion}
In this section, we first evaluate the performance of the geometric foundation model and demonstrate how the guided diffusion model generates heat-sink geometries conditioned on a specified temperature constraint while minimizing pressure drop. We illustrate the evolution of these geometries across the timesteps of the guidance process, followed by a demonstration of the optimized designs and their statistical distributions at selected temperature thresholds. Finally, we compare the results against the CMA-ES optimization baseline and provide full CFD validation of the optimized configurations.

\begin{figure}[h]
  \centering
  \begin{subfigure}[b]{0.48\textwidth}
    \centering
    \includegraphics[width=\textwidth]{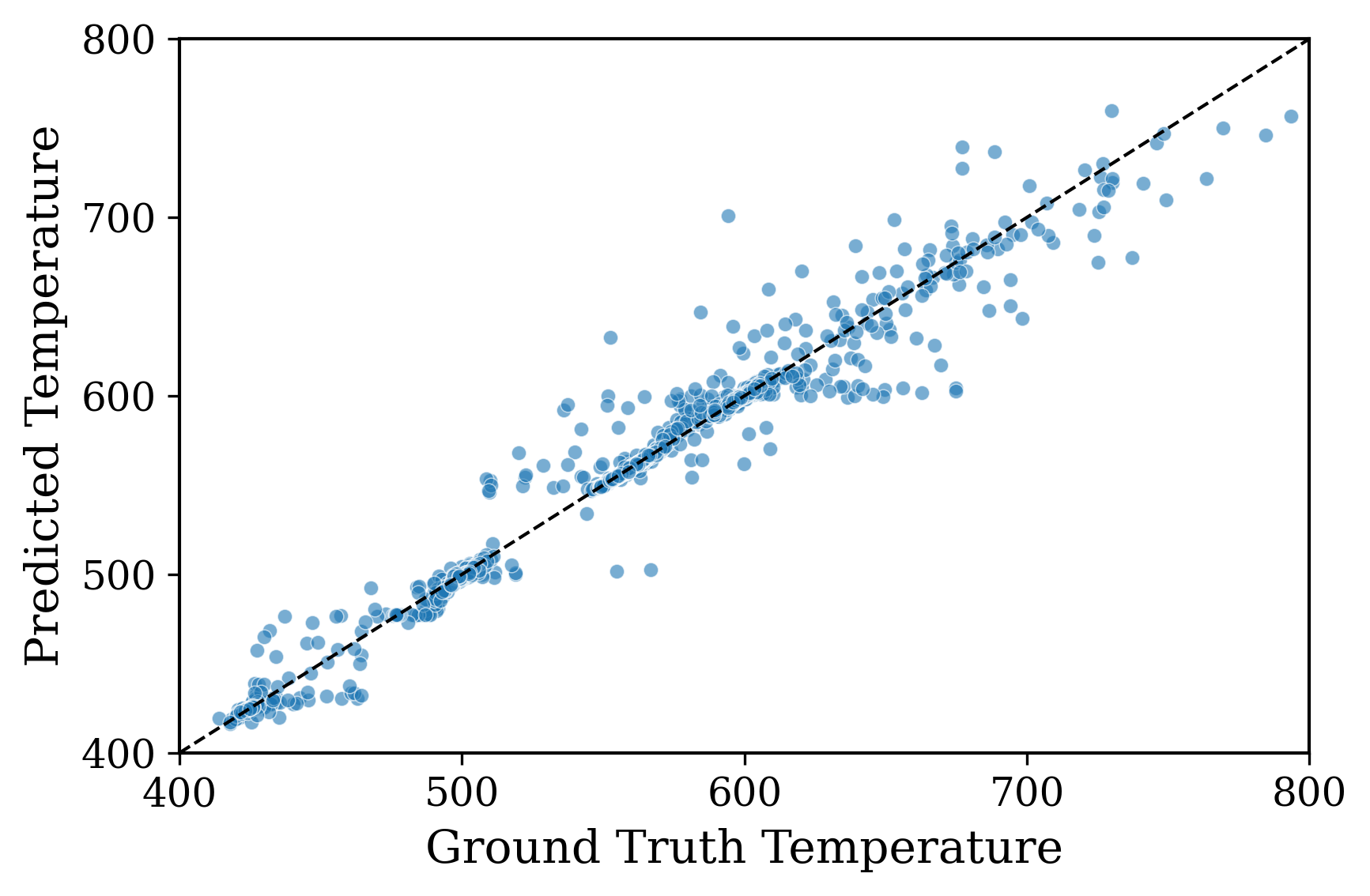}
    \caption{}
    \label{fig:a}
  \end{subfigure}
  \hfill
  \begin{subfigure}[b]{0.48\textwidth}
    \centering
    \includegraphics[width=\textwidth]{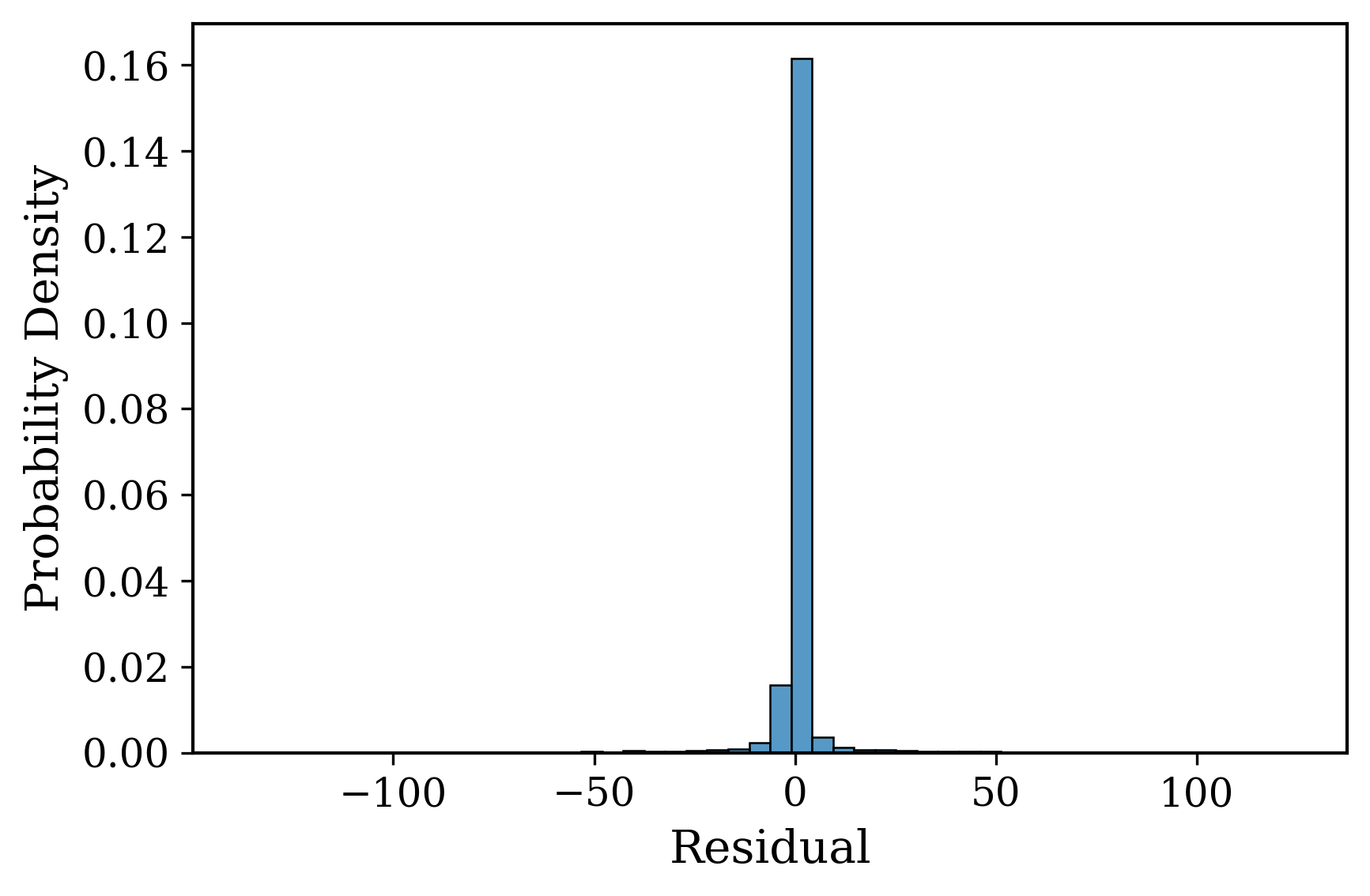}
    \caption{}
    \label{fig:Tpred}
  \end{subfigure}
  
  \caption{Surrogate model performance for average surface temperature prediction: (a) comparison of predicted and ground truth values and (b) residual distribution, defined as prediction minus actual value.}
  \label{fig:TregpredvsTrue}
\end{figure}

\begin{figure}[h]
  \centering
  \begin{subfigure}[b]{0.48\textwidth}
    \centering
    \includegraphics[width=\textwidth]{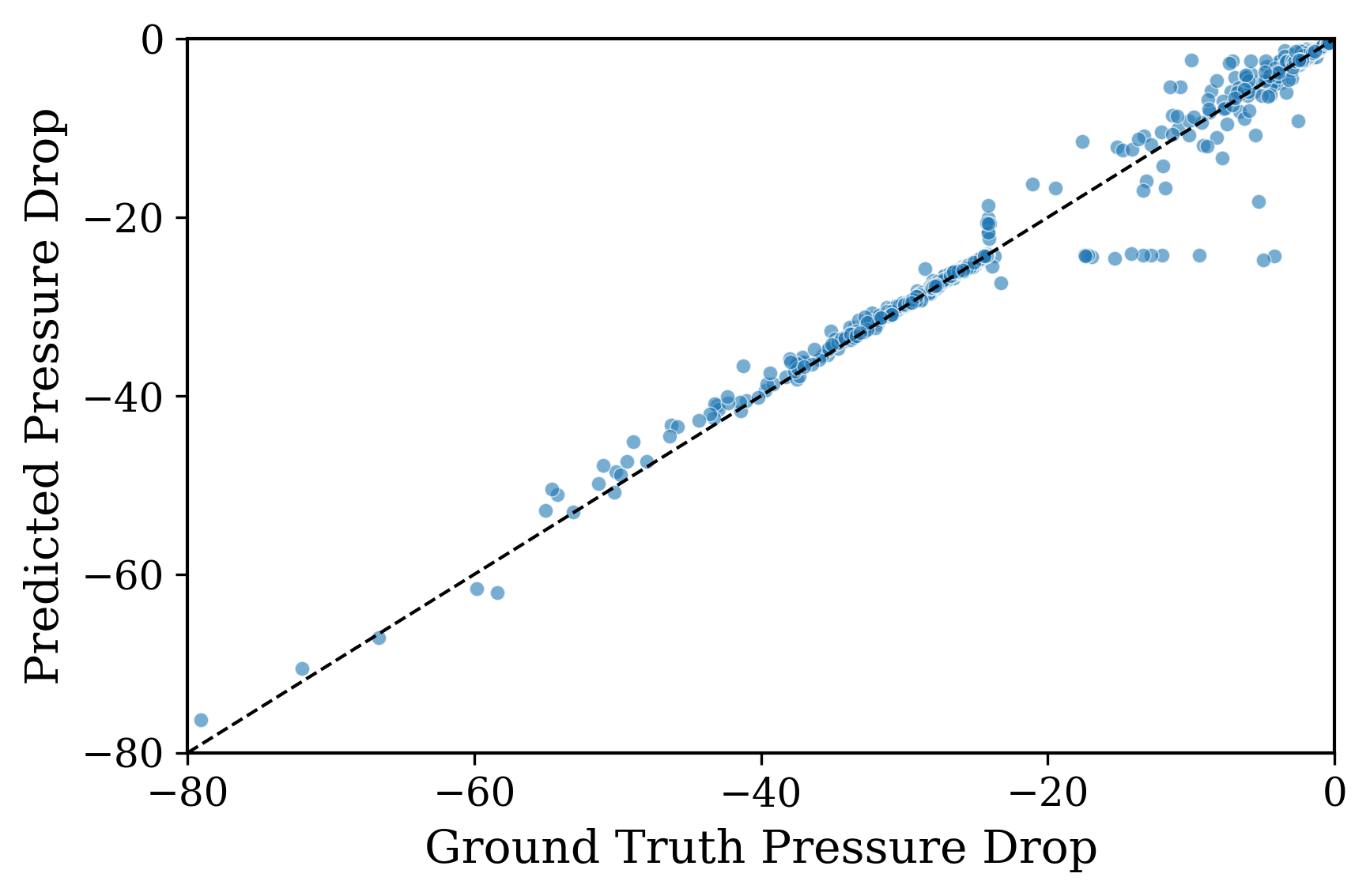}
    \caption{}
    \label{fig:a}
  \end{subfigure}
  \hfill
  \begin{subfigure}[b]{0.48\textwidth}
    \centering
    \includegraphics[width=\textwidth]{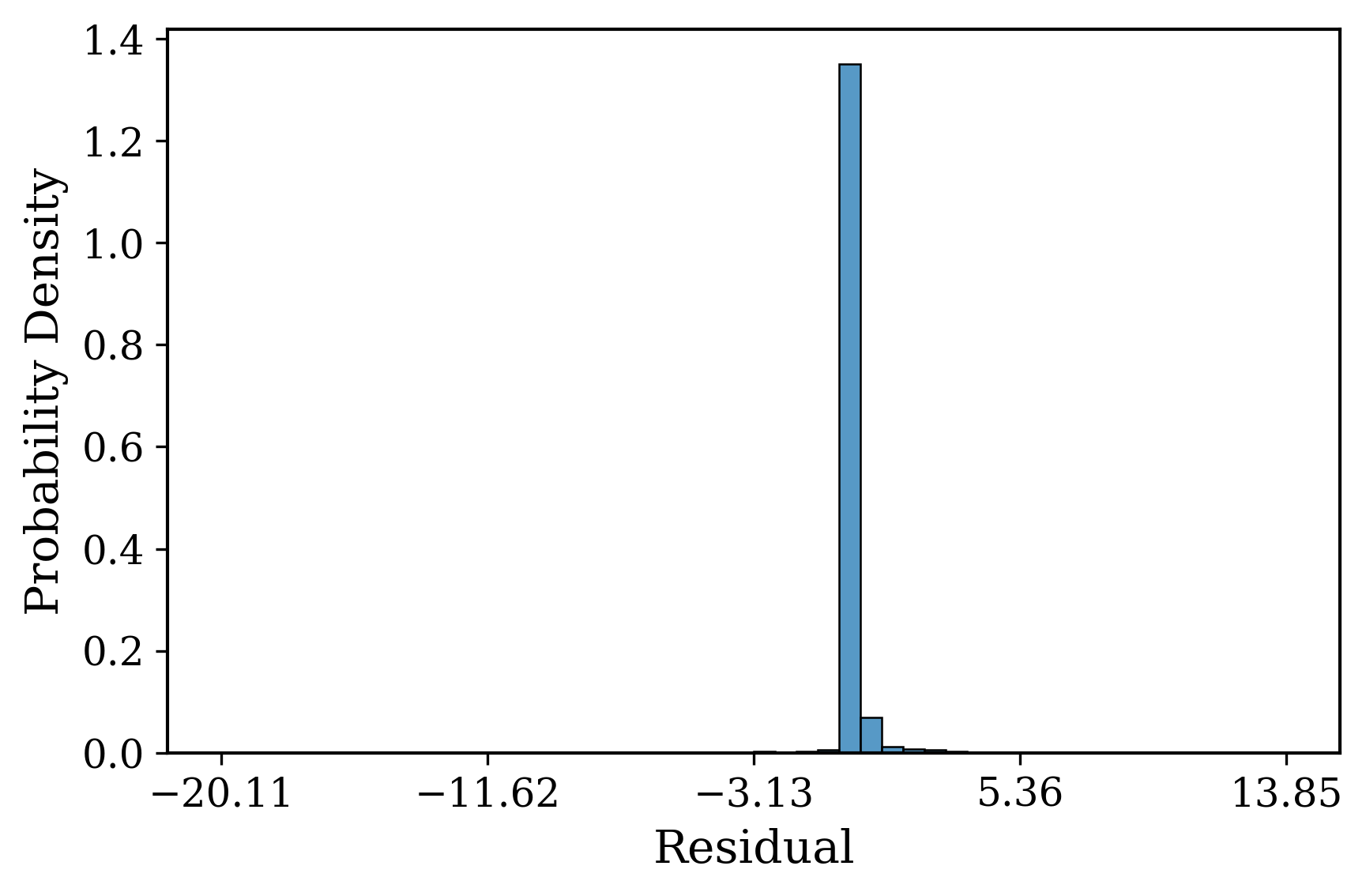}
    \caption{}
    \label{fig:b}
  \end{subfigure}  
  \caption{Performance of the surrogate model for pressure prediction: (a) predicted versus ground-truth values, and (b) residuals computed as the difference between predicted and actual values.}
  \label{fig:Ppred}
\end{figure}

Since the framework largely depends on the performance of our geometric model, two different models were trained to predict the pressure drop and surface temperature. Although the surface temperature model was more challenging to train compared to the overall heat transfer and pressure drop, the final model, hyperparameterized on the Weights \& Biases platform using random search, achieved a high value of $R^2 = 0.985$ on the test data, which the model had not previously seen. The predicted and actual values, as well as the histogram of the difference between them are shown in Fig.~\ref{fig:TregpredvsTrue}. The implementation details of the pressure regression model are explained in Fig. \ref{fig:pressure_resmlp_layers} of \ref{app:modelap}, and the performance of the pressure model with an $R^2$ value of 0.989 for the unseen test data is shown in Fig. \ref{fig:Ppred}.

\begin{figure}[h!]
    \centering
    \begin{subfigure}{0.4\textwidth}
        \includegraphics[width=\linewidth]{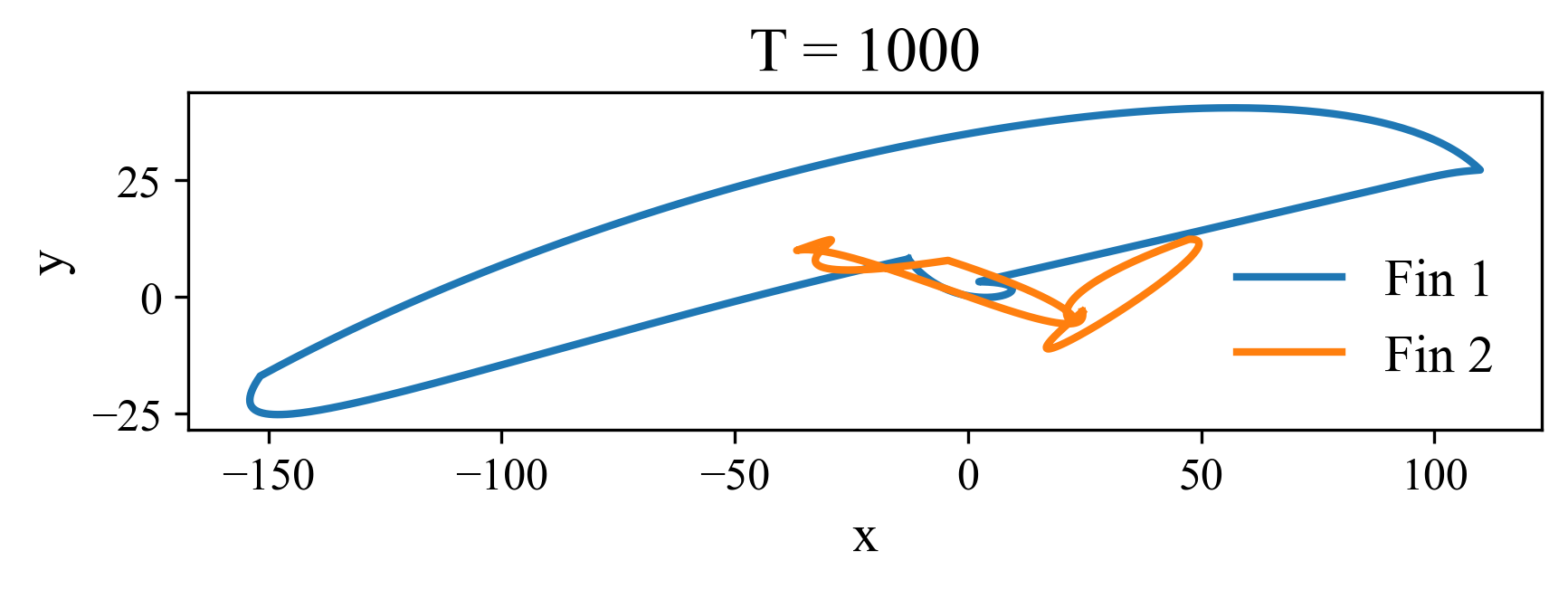}
        \caption{Noisy state with pure Gaussian noise }
    \end{subfigure}
    \begin{subfigure}{0.4\textwidth}
        \includegraphics[width=\linewidth]{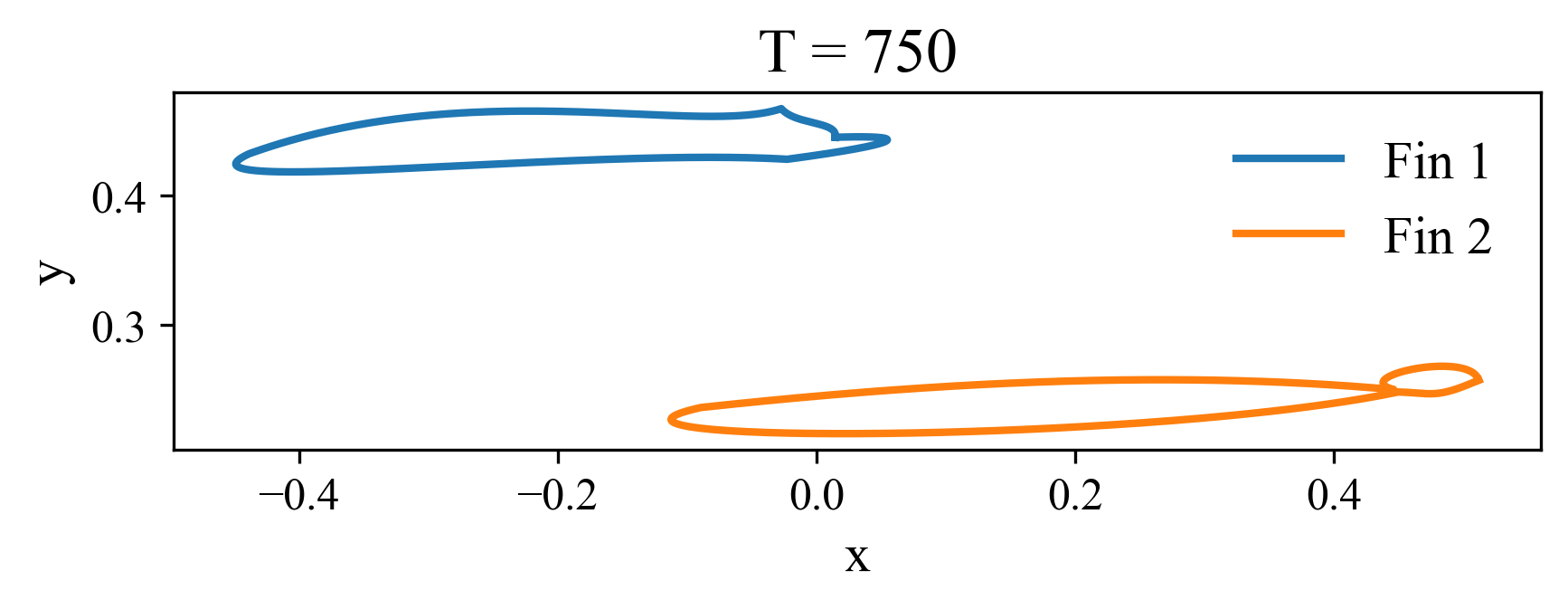}
        \caption{Initial phase of denoising at $ t = 750$}
    \end{subfigure}
    \begin{subfigure}{0.4\textwidth}
        \includegraphics[width=\linewidth]{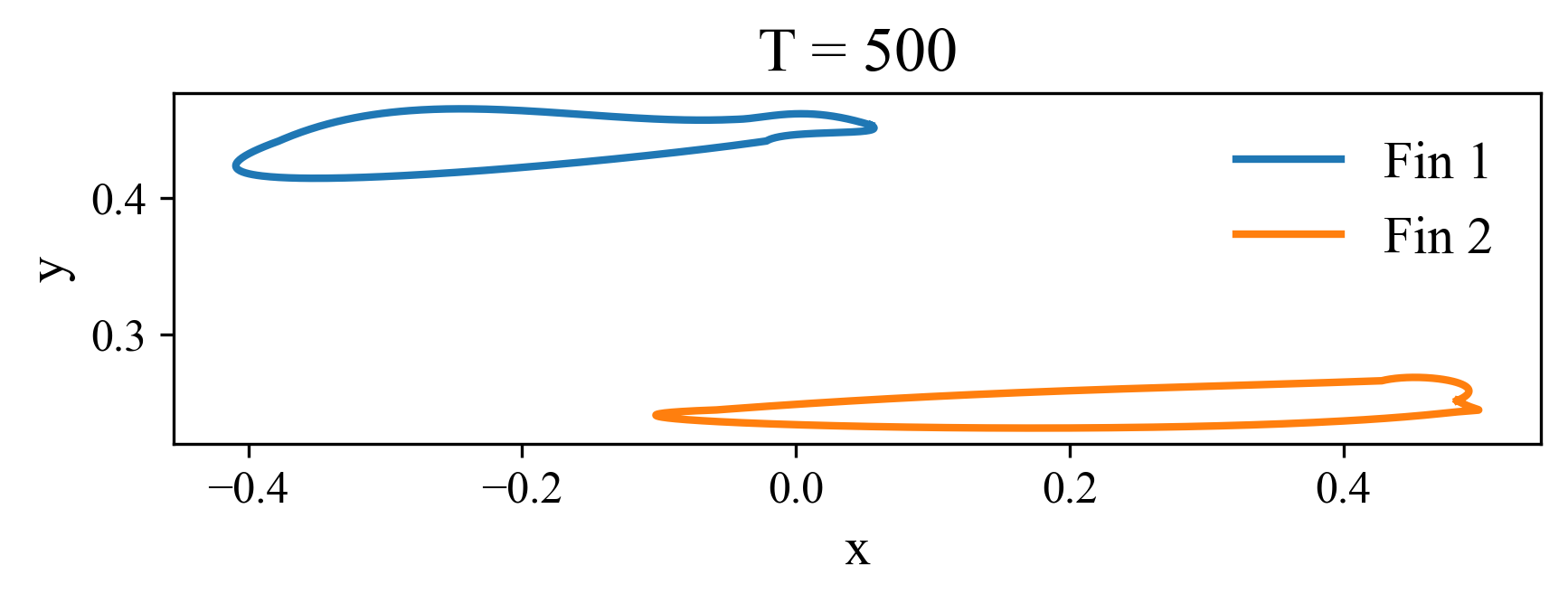}
        \caption{Intermediate state in the reverse process}
    \end{subfigure}
    \begin{subfigure}{0.4\textwidth}
        \includegraphics[width=\linewidth]{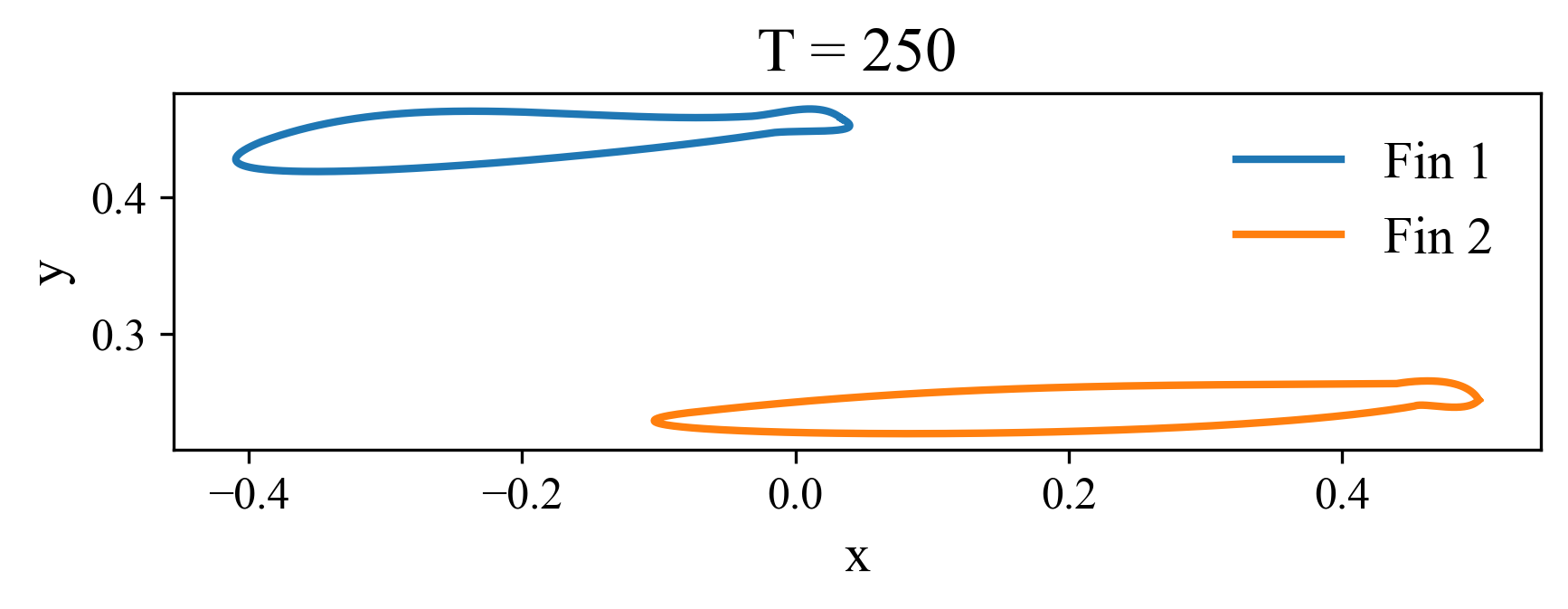}
        \caption{Final phase of denoising at $ T = 250$}
    \end{subfigure}
    \begin{subfigure}{0.4\textwidth}
        \includegraphics[width=\linewidth]{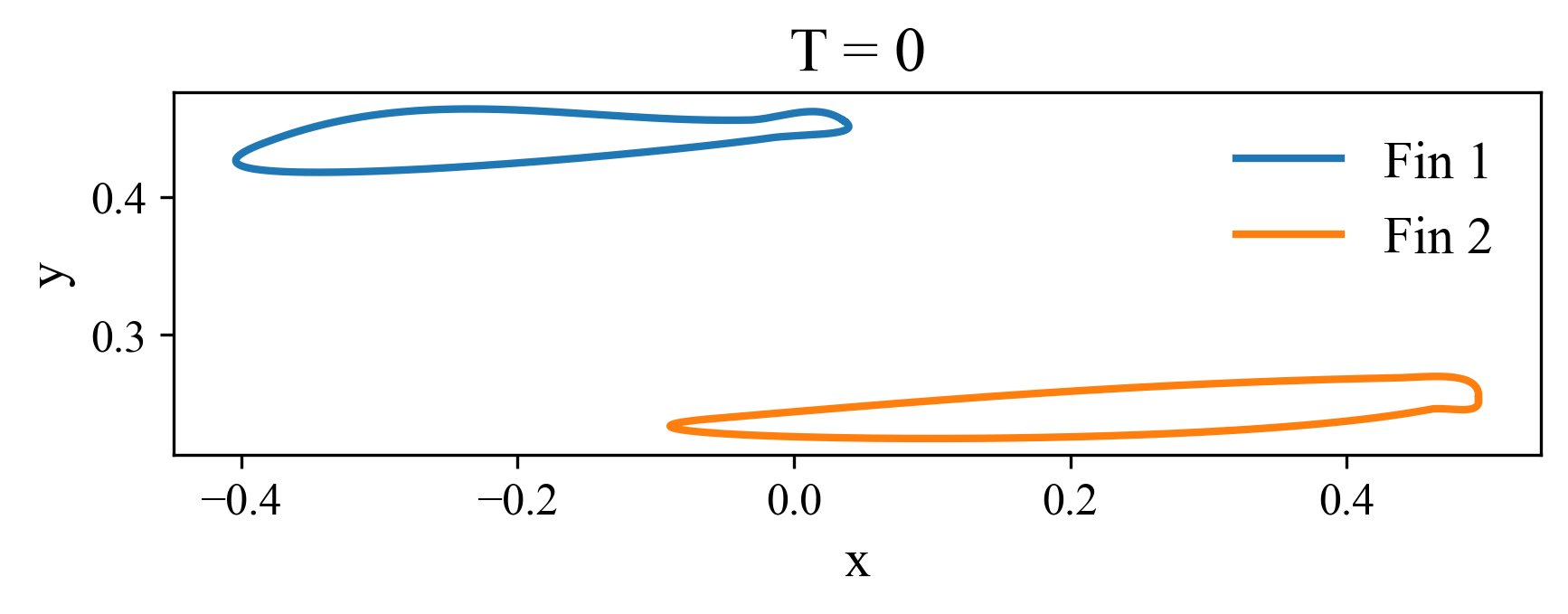}
        \caption{Final shape}
    \end{subfigure}
    \caption{Evolution of the noise reduction from the fin shapes across the denoising process to produce designs with reduced pressure drop and surface tmperature of \(550~\text{K}\)}
    \label{fig:evol}
\end{figure}

Using the gradients from these high-accuracy geometric models, the guidance process progressively reduces noise from an initial pure noise state toward a fully optimized shape. Fig.~\ref{fig:evol} illustrates the evolution of the generated geometries from \(T = 1000\), corresponding to pure noise, down to \(x_0\) at \(T = 0\), representing the optimized design at a temperature of \(550~\text{K}\). At \(T = 1000\), the fin geometries are completely random and intersect with each other. Unlike image generation, where noise is added to pixel values within a fixed spatial domain, in HeatGen the noise is applied directly to the geometry vector, whose values define the spatial domain itself. Therefore, the domain at \(T = 1000\) is not directly comparable to other timesteps due to the unstructured nature of the noise. By \(T = 750\), the shapes begin to form slender bodies resembling those in the training data, but self-intersections and acute angles can still be observed. At \(T = 500\), the noise responsible for the self-intersections is mainly removed, but sharp edges and acute corners that contribute to high pressure drops still exist. At \(T = 250\), no self-intersections or unexpected edges persist, and the shapes become smoother, though not yet fully optimized to minimize pressure drop and prevent hotspots. Finally, at \(T = 0\),  the noise is completely eliminated, smooth continuous contours emerge and both both thermal and hydraulic performance objectives are satisfied, confirming the successful convergence of the guided diffusion process.

\begin{figure}[htbp]
  \centering
  \begin{subfigure}[b]{0.48\textwidth}
    \centering
    \includegraphics[width=\textwidth]{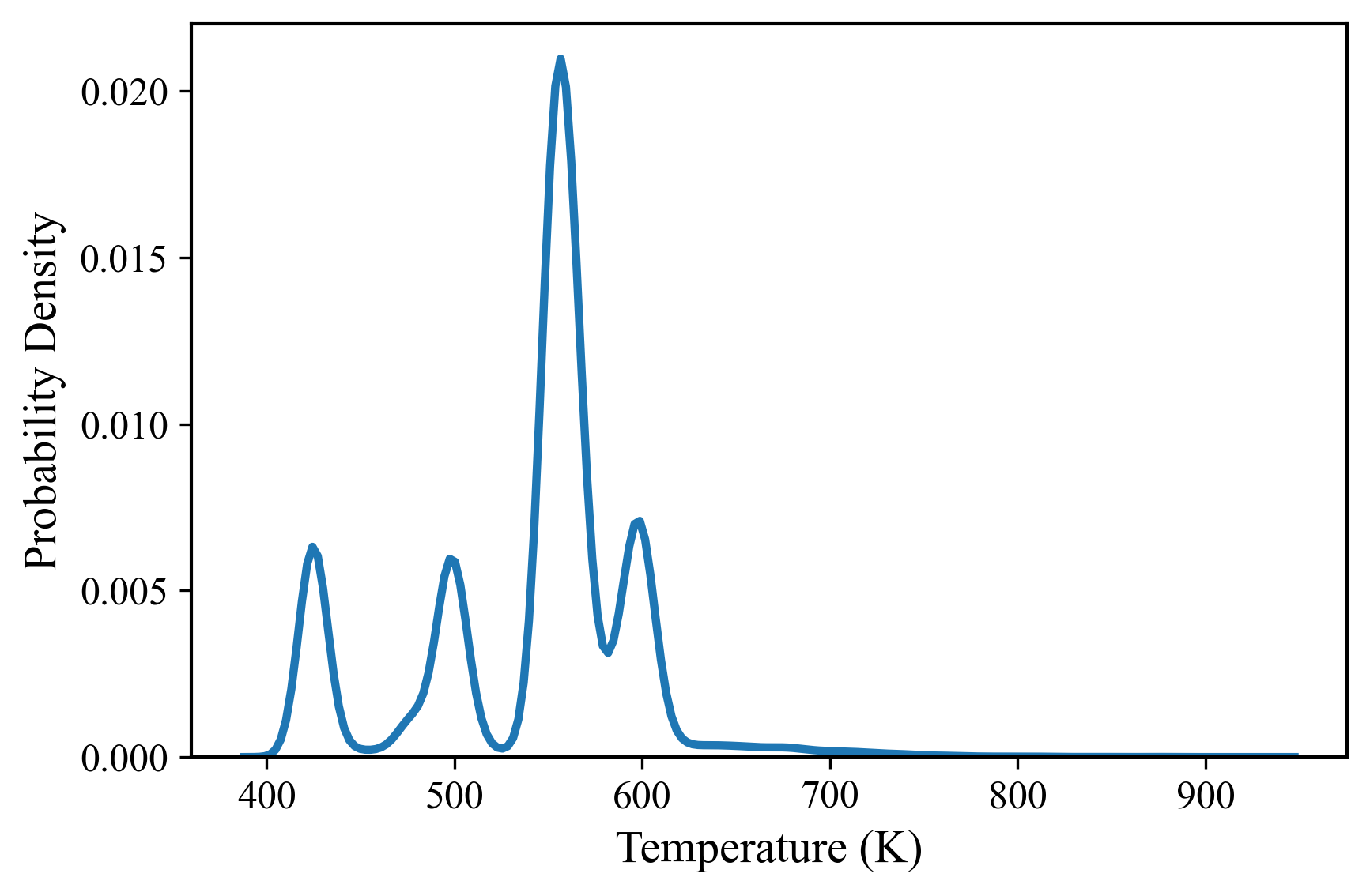}
    \caption{Temperature}
    \label{fig:a}
  \end{subfigure}
  \hfill
  \begin{subfigure}[b]{0.48\textwidth}
    \centering
    \includegraphics[width=\textwidth]{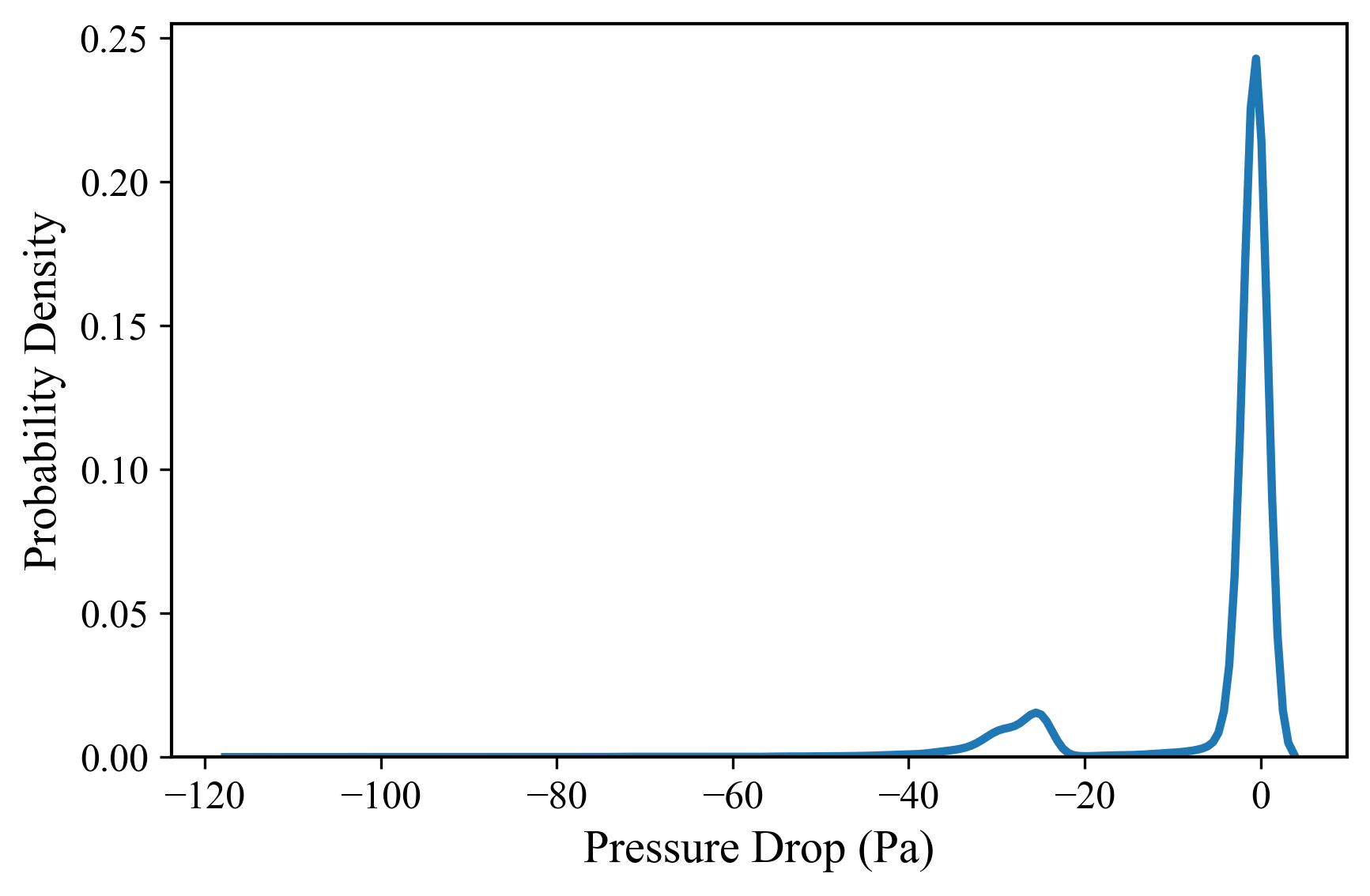}
    \caption{Pressure}
    \label{fig:b}
  \end{subfigure}
  
  \caption{Probability density functions of the training data for (a) temperature and (b) pressure}
  \label{fig:trainingkde}
\end{figure}

\begin{figure}[htbp]
  \centering
  \begin{subfigure}[b]{0.48\textwidth}
    \centering
    \includegraphics[width=\textwidth]{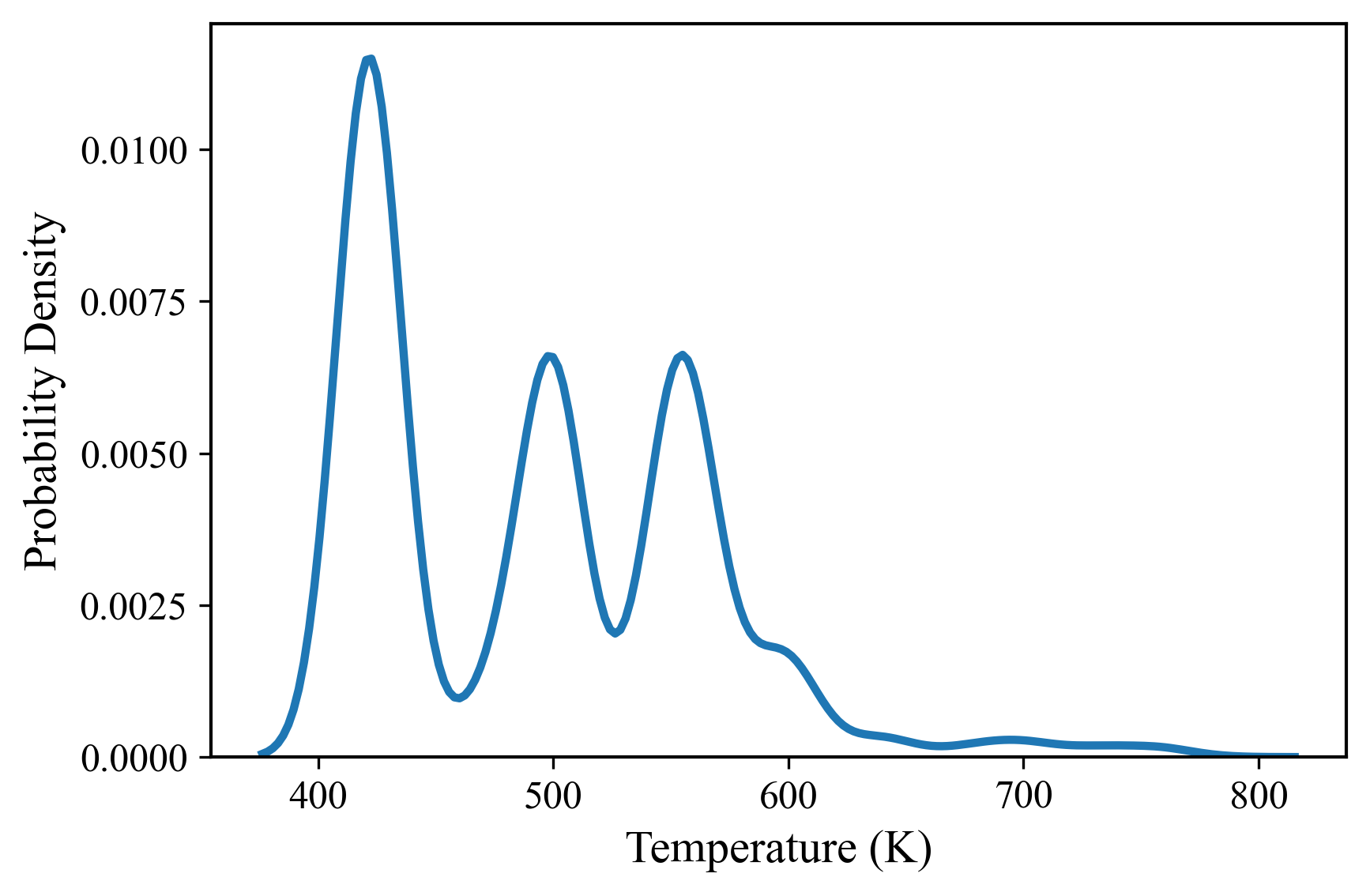}
    \caption{Temperature density function conditioned on $ 425 K$}
    \label{fig:a}
  \end{subfigure}
  \hfill
  \begin{subfigure}[b]{0.48\textwidth}
    \centering
    \includegraphics[width=\textwidth]{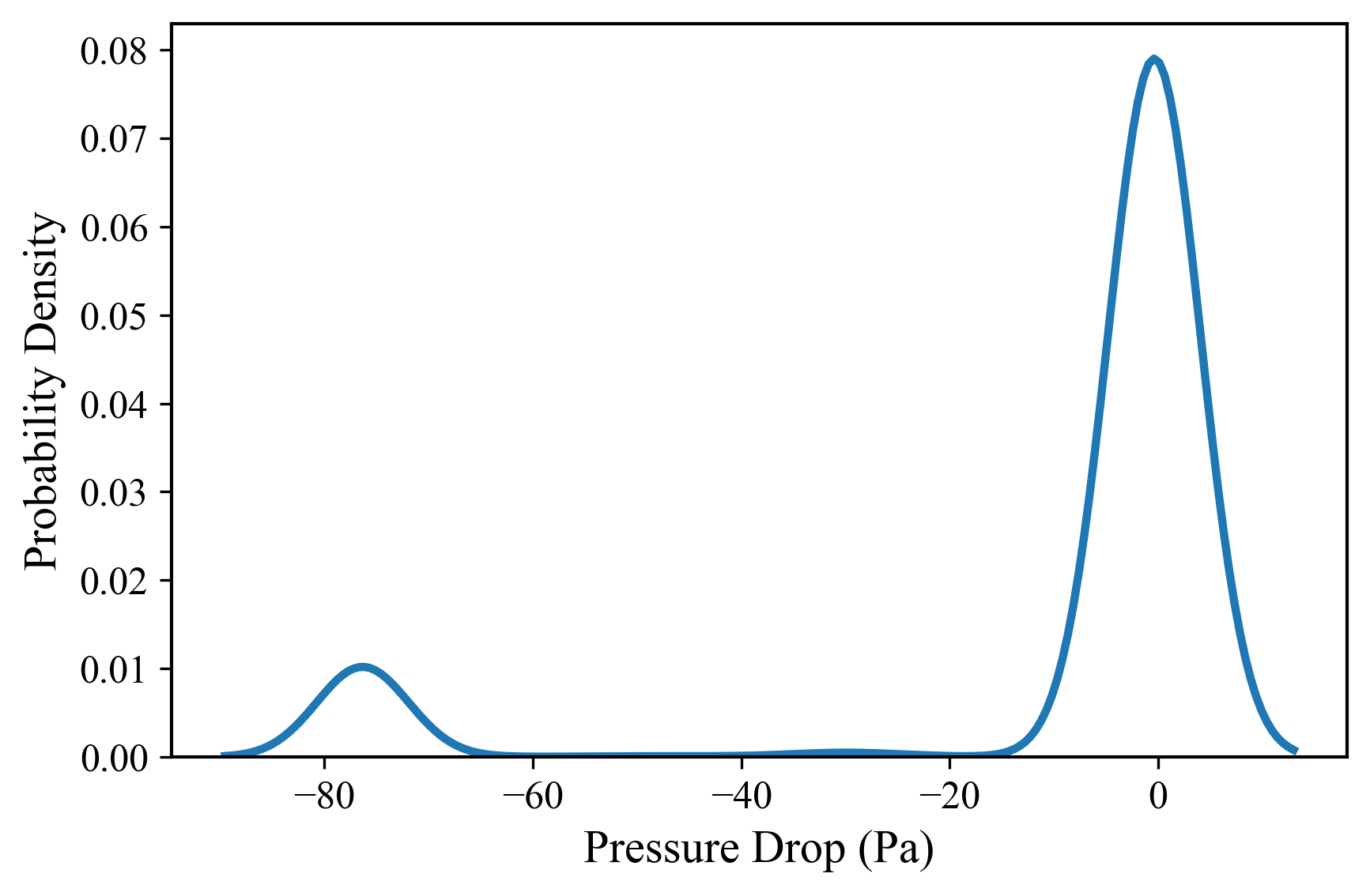}
    \caption{Pressure density function conditioned on $ 425 K$}
    \label{fig:b}
  \end{subfigure}
  
  \caption{Probability density functions of (a) temperature and (b) pressure obtained from the guided diffusion process conditioned on a maximum temperature of \(425~\text{K}\). The peak of the guided distribution appears around \(425~\text{K}\)(guidance scales of \(\eta = 0.01, \lambda_p = 0.4, \lambda_T = 0.4\)).}
  \label{fig:425kde}
\end{figure}

\begin{figure}[h]
  \centering
  \begin{subfigure}[b]{0.48\textwidth}
    \centering
    \includegraphics[width=\textwidth]{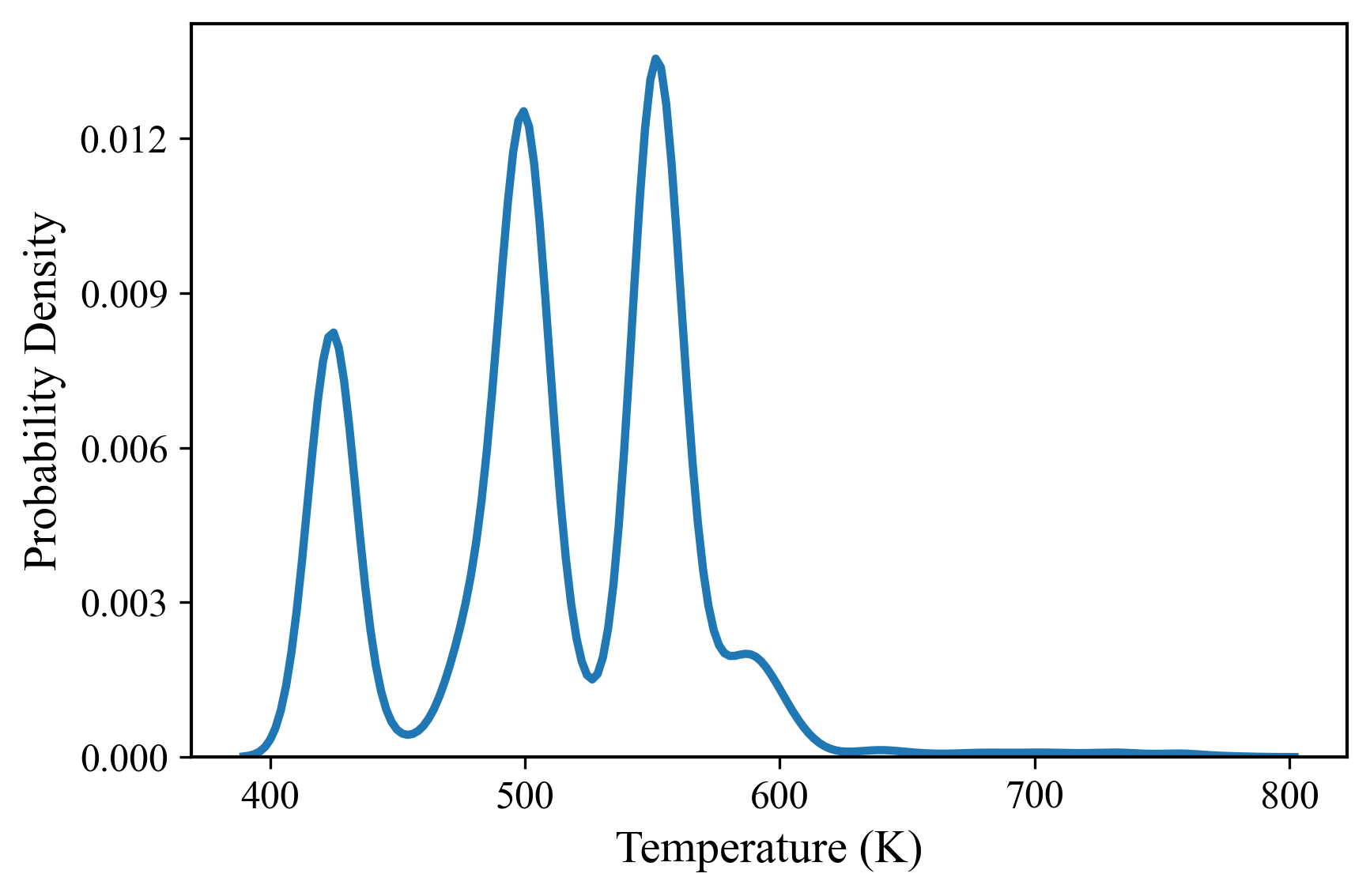}
    \caption{Temperature density function conditioned on $ 500 K$}
    \label{fig:a}
  \end{subfigure}
  \hfill
  \begin{subfigure}[b]{0.48\textwidth}
    \centering
    \includegraphics[width=\textwidth]{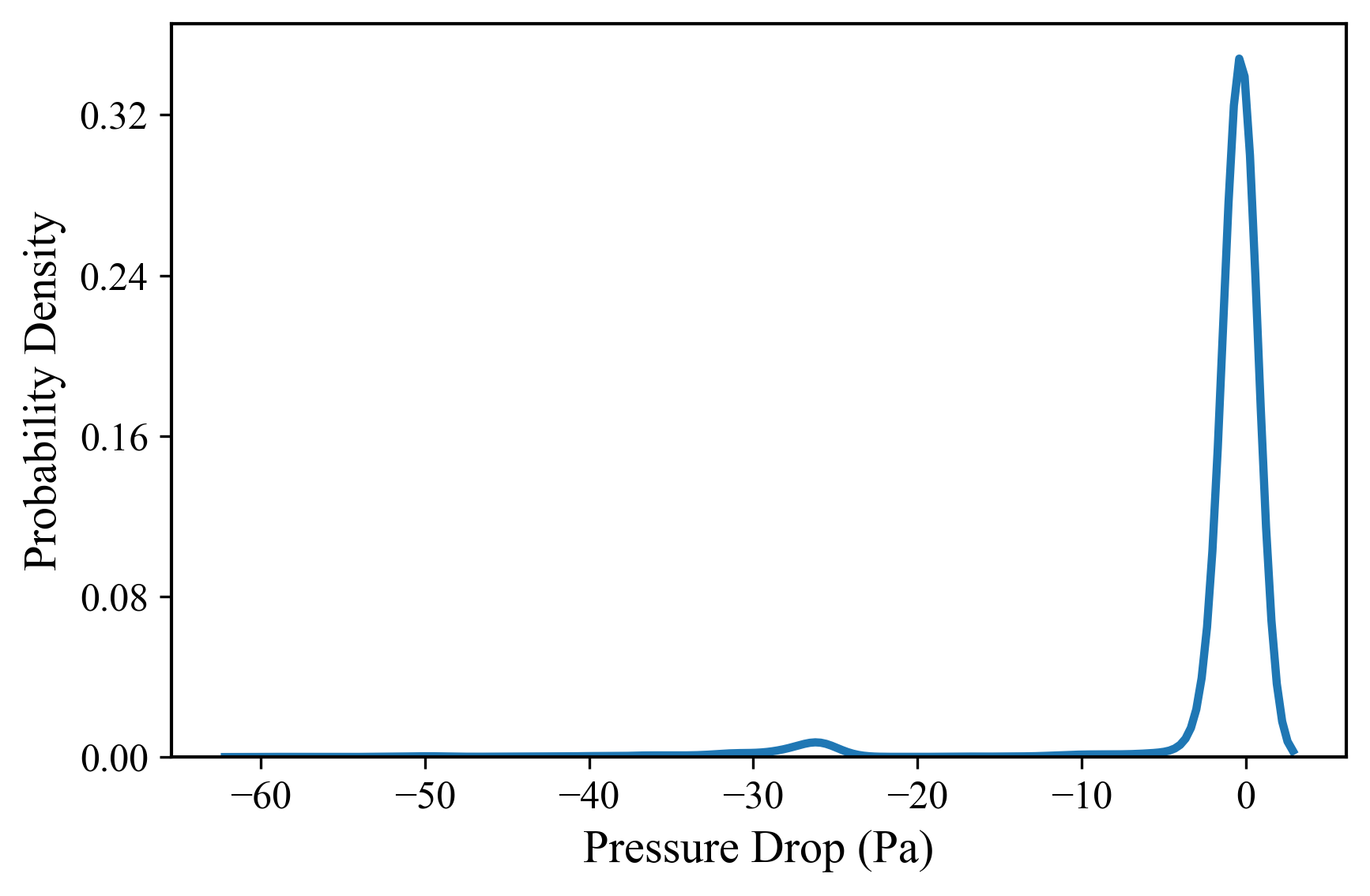}
    \caption{Pressure density function conditioned on $ 500 K$}
    \label{fig:b}
  \end{subfigure}
  
  \caption{Probability density functions of (a) temperature and (b) pressure obtained from the guided diffusion process conditioned on maximum temperature of \(500~\text{K}\) (guidance scales of \(\eta = 0.01, \lambda_p = 0.4, \lambda_T = 0.4\)).}
  \label{fig:kde500}
\end{figure}

Fig.~\ref{fig:trainingkde} shows the probability distributions of temperature and pressure in the training dataset. The data exhibit a multimodal behavior, with multiple peaks observed particularly in the temperature distribution.
Fig.~\ref{fig:425kde} shows the probability distributions of pressure and temperature conditioned on a maximum temperature of \(425~\text{K}\). The peak of the guided distribution occurs around \(425~\text{K}\), whereas the peak of the training data distribution was located near \(575~\text{K}\).
Fig.~\ref{fig:kde500} shows the probability density functions of temperature and pressure when the guidance was conditioned on a maximum temperature of \(500~\text{K}\). The parameters of the guidance process for both results shown in Fig.~\ref{fig:425kde} and Fig.~\ref{fig:kde500} are \(\eta = 0.01, \lambda_p = 0.4, \lambda_T = 0.4\). Although the peak of the generated samples does not occur exactly at \(500~\text{K}\), the density of samples near this temperature increases substantially compared to the training data. This indicates that the guidance process effectively shifts the generated designs toward the specified hotspot limit while simultaneously reducing pressure. The generated designs from these inference constraints are over 98\% feasible, which means that the surrogate prediction is within 5\% of the pseudo-3D model results.

\begin{figure}[h]
  \centering
  \begin{subfigure}[b]{0.48\textwidth}
    \centering
    \includegraphics[width=\textwidth]{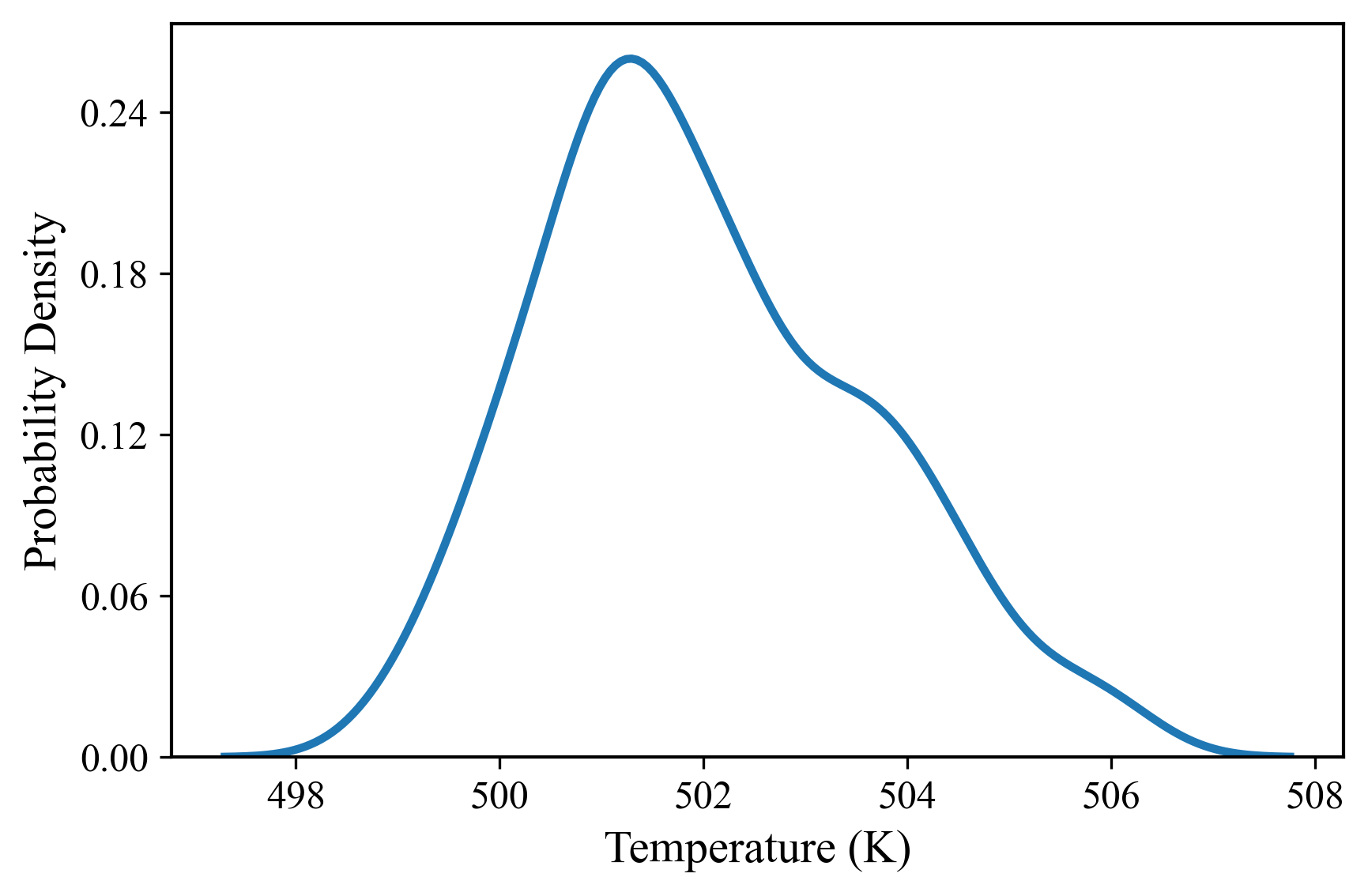}
    \caption{Temperature density function conditioned on $ 500 K$}
    \label{fig:a}
  \end{subfigure}
  \hfill
  \begin{subfigure}[b]{0.48\textwidth}
    \centering
    \includegraphics[width=\textwidth]{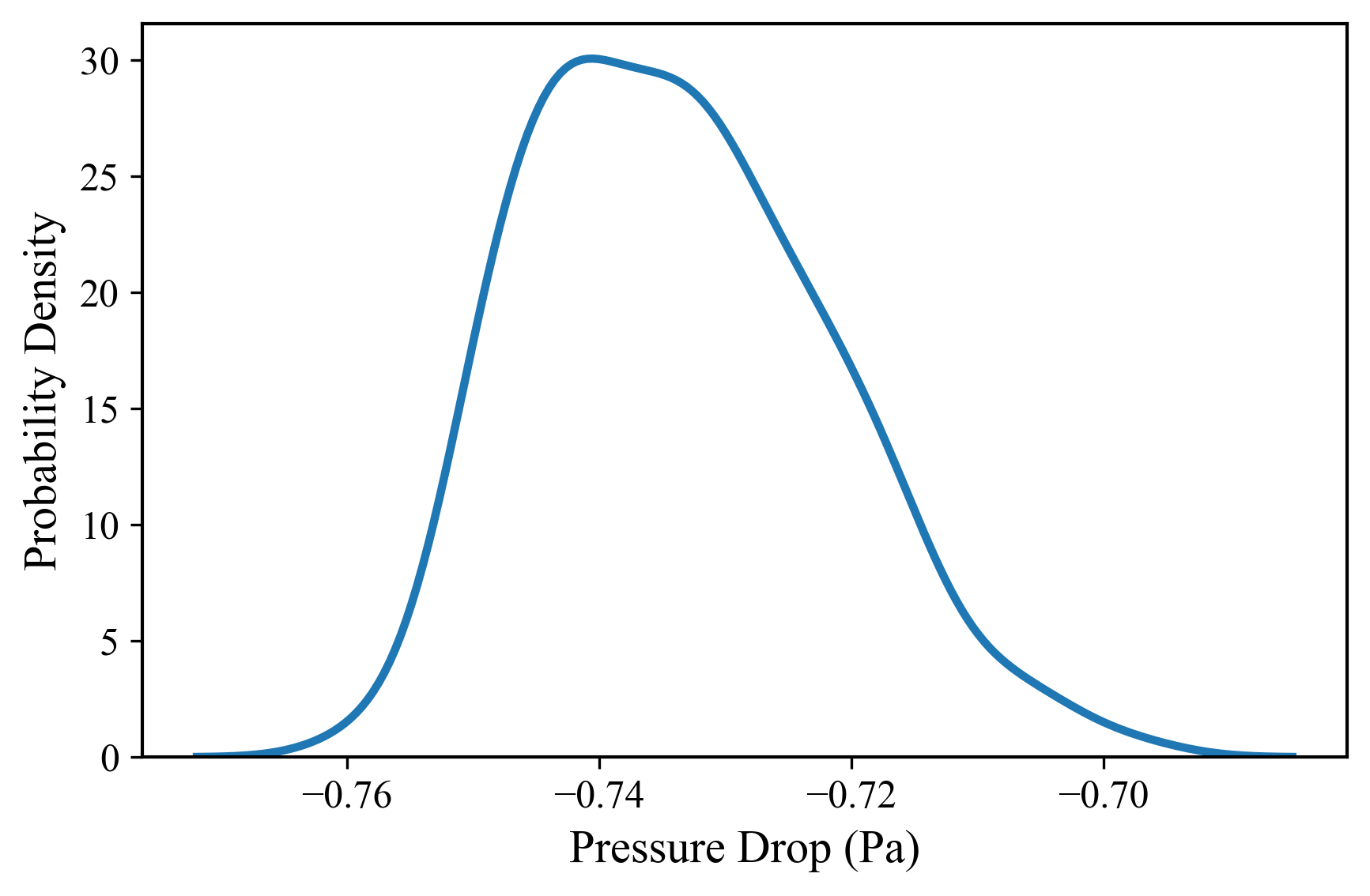}
    \caption{Pressure density function conditioned on $ 500 K$}
    \label{fig:b}
  \end{subfigure}
  
    \caption{Probability density functions of (a) temperature and (b) pressure obtained from the largely guided diffusion process conditioned on a maximum temperature of \(500~\text{K}\); the feasibility rate is reduced to 74\%.}
  \label{fig:kde500highlyguided}
\end{figure}

Changing the scale of the guidance might result in a larger shift in the data distribution, as shown in Fig.~\ref{fig:kde500highlyguided} with parameters of \(\eta = 0.3, \lambda_p = 0.3, \lambda_T = 0.5\), but the feasibility rate of the generated designs is reduced to 74\% since there is no feasibility guidance. 
To assess the performance of optimized designs under the constraints imposed relative to traditional optimization methods, Table~\ref{tab:compareheatgencmaes} presents a comparison in temperatures ranging from \(400~\text{K}\) to \(550~\text{K}\). For a fair comparison, the CMA-ES algorithm was allowed the same number of iterations as HeatGen, but in practice, CMA-ES shows little to no improvement beyond 500 iterations. Across nearly the entire temperature range, HeatGen outperforms CMA-ES in minimizing pressure drop, achieving up to 10\% reduction compared to the CMA-ES results.

\begin{table}[h]
\centering
\caption{Elite designs with minimum pressure drops across different maximum surface temperature ranges for CMA-ES and HeatGen.}
\label{tab:compareheatgencmaes}
\begin{tabular}{lccccccc}
\toprule
 Method & 400 & 425 & 450 & 475 & 500 & 550  \\
\midrule
HeatGen & 38.136 & 23.517 & 5.261 & 2.298 & 0.746 & 0.464  \\
CMA-ES & 38.057 & 24.270 & 5.647 & 2.529 & 0.789 & 0.473  \\
Improvement ($\%$) & -- & 3.1 & 6.84 & 9.1 & 5.8 & 1.9 \\
\bottomrule
\end{tabular}
\end{table}

\begin{figure}[h]
    \centering
    \includegraphics[width=0.7\linewidth]{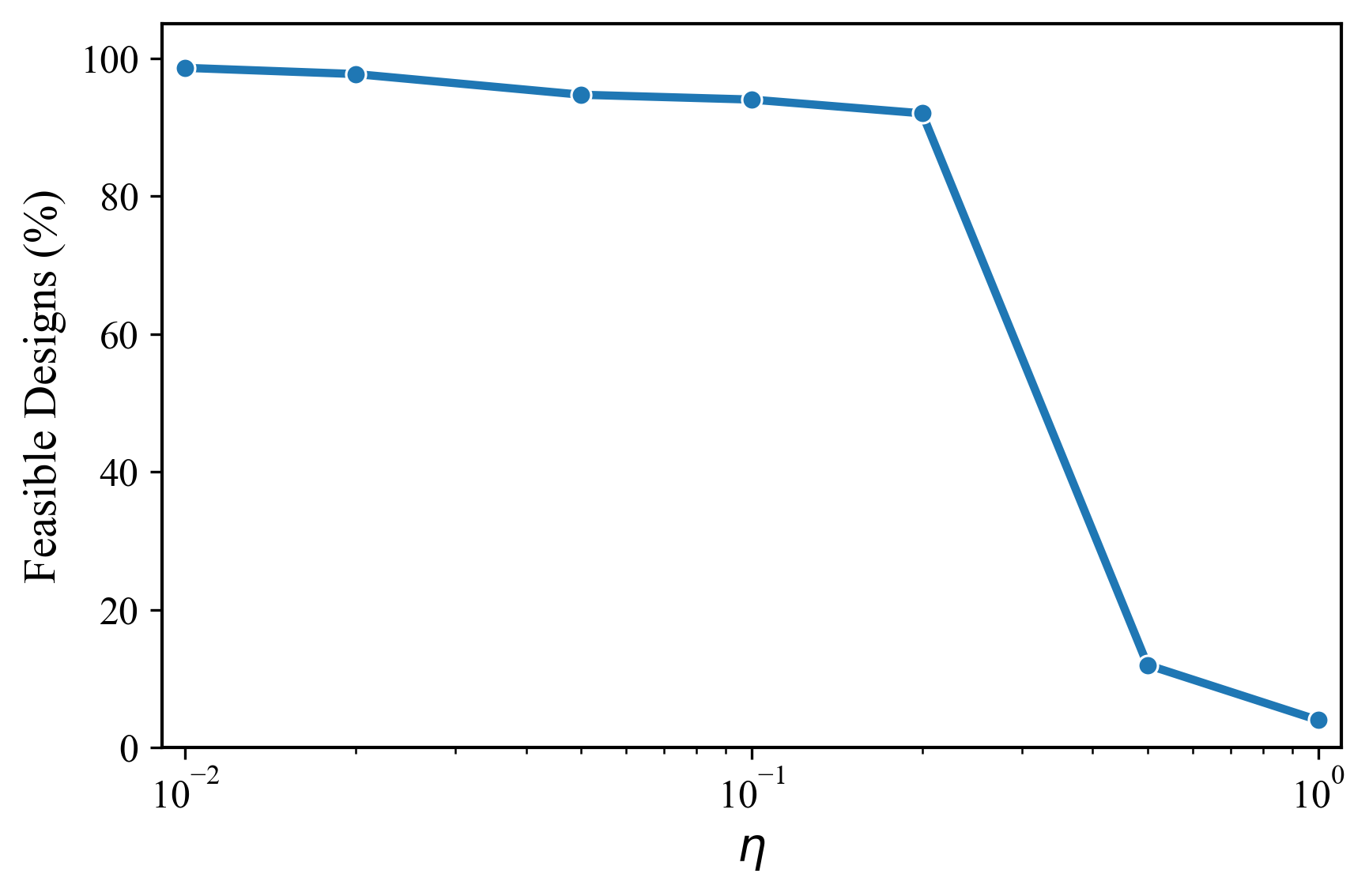}
    \caption{Percentage of feasible design at different guidance impact: higher $\eta$ means higher guidance impact.}
    \label{fig:feasibilityeta}
\end{figure}

To evaluate the impact of guidance on the number of invalid designs generated by HeatGen, we analyze the effect of the parameter \(\eta\), which controls the influence of the gradient on the modification of the distribution. Fig.~\ref{fig:feasibilityeta} illustrates that decreasing \(\eta\) to 0.01, while keeping the pressure and temperature surrogate weights fixed at \(\lambda_{p} = 0.4\) and \(\lambda_{T} = 0.4\), results in a higher percentage of feasible designs. This behavior could change if an additional gradient term from a classifier that determines the feasibility of the design were introduced to guide the generation process by adding a third gradient term with its corresponding weight. However, since a feasibility classifier is not available and the focus here is solely on performance-based guidance, the feasible designs are plotted based on the gradient weight of the performance and constraint objectives.

\begin{figure}[h]
    \centering
    \includegraphics[width=0.7\linewidth]{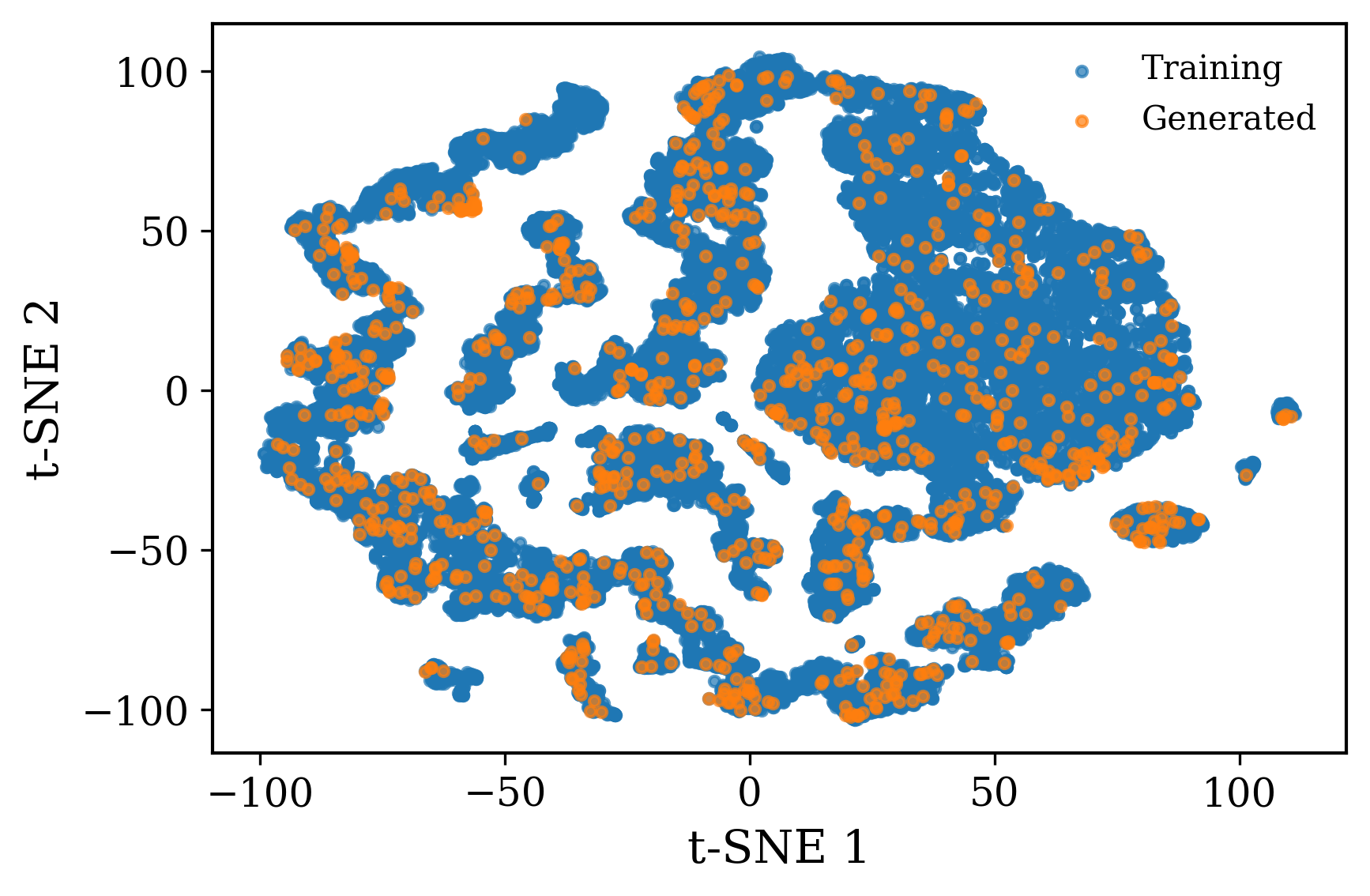}
    \caption{t-SNE of the generated samples from the DDPM without guidance or constraints}
    \label{fig:tsepuresample}
\end{figure}

\begin{figure}[H]
    \centering
    \includegraphics[width=0.7\linewidth]{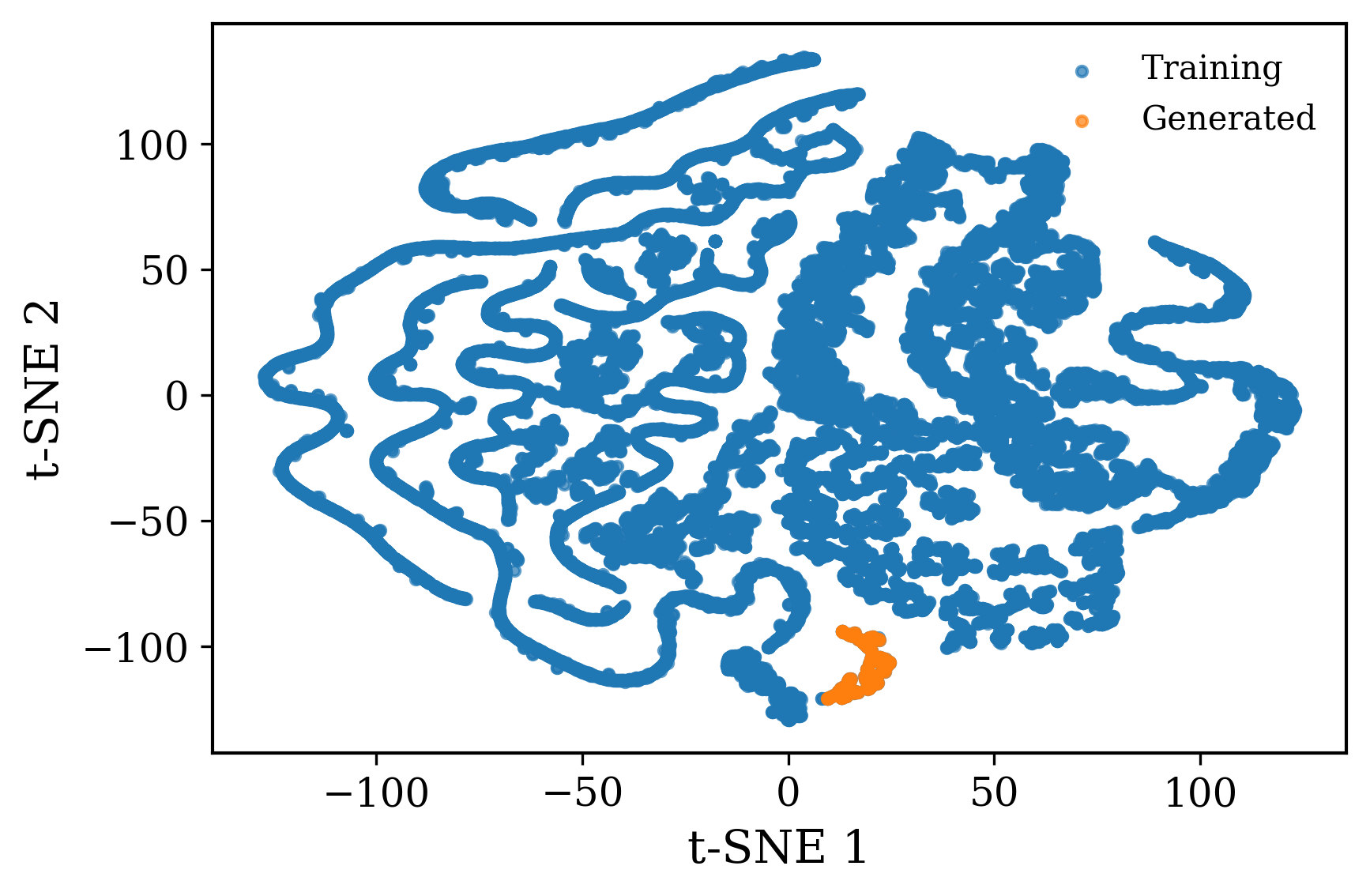}
    \caption{t-SNE of the highly guided design generation conditioned on a maximum surface temperature of \(500~\text{K}\) and pressure minimization}
    \label{fig:tseguidedsample}
\end{figure}

Training the DDPM without applying gradient-based guidance generates designs that are similar to the initial samples. To illustrate the data coverage, we use t-SNE, a visualization technique that represents high-dimensional data in a lower-dimensional space~\cite{maaten2008visualizing}. Fig.~\ref{fig:tsepuresample} shows the t-SNE projection of the sample heat sinks generated without any guidance, where the generated samples in red cover the entire training data space shown in blue. The highly guided feasible samples that generate heat sinks with a temperature constraint of \(500~\text{K}\) and minimum pressure are shown in Fig.~\ref{fig:tseguidedsample}. It can be seen that guided generation shifts the samples toward a specific direction in the reduced-dimensional t-SNE visualization.

\begin{figure}[H]
    \centering
    \begin{subfigure}{1\textwidth}
        \includegraphics[width=0.95\linewidth]{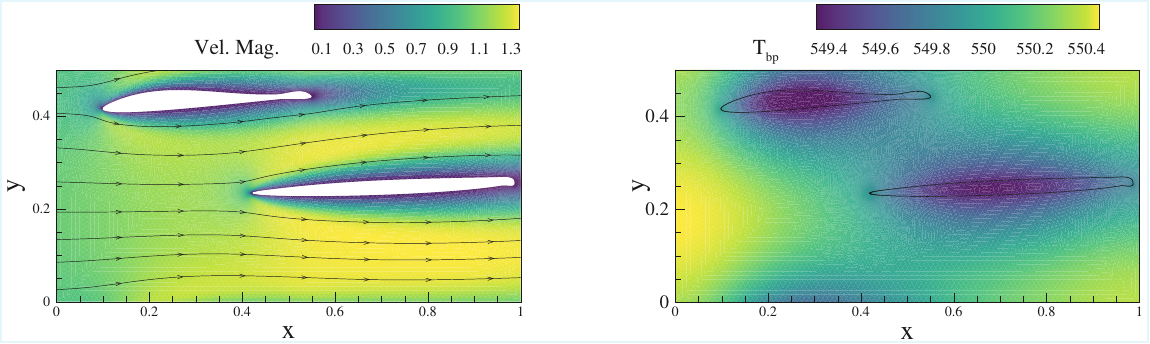}
        \caption{}
    \end{subfigure}

        \begin{subfigure}{1\textwidth}
        \includegraphics[width=0.95\linewidth]{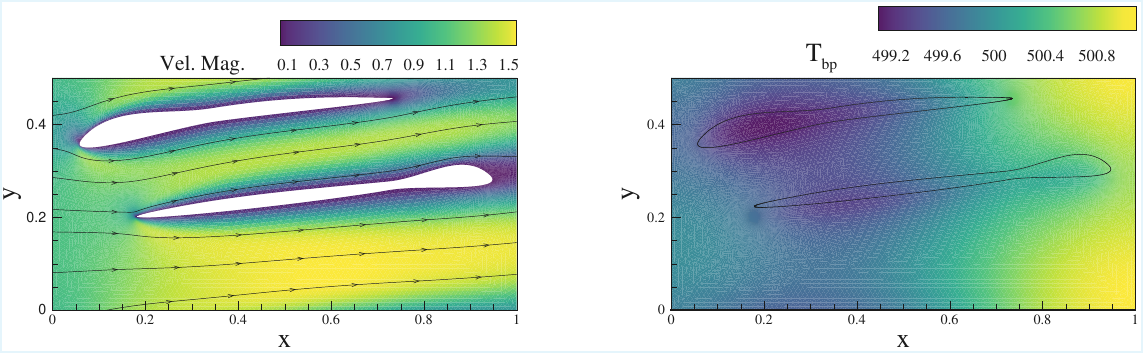}
        \caption{ }
    \end{subfigure}

            \begin{subfigure}{1\textwidth}
        \includegraphics[width=0.95\linewidth]{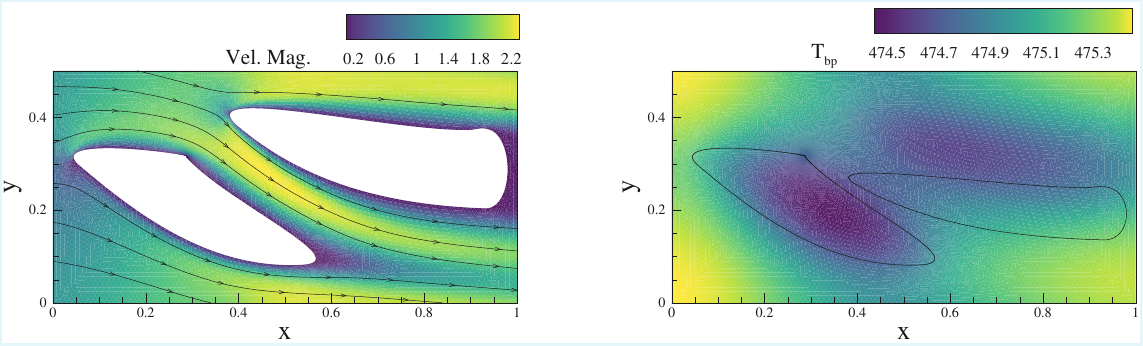}
        \caption{}
    \end{subfigure}

    \caption{Velocity field distribution (left) and temperature field distribution (right) for three optimized fin configurations generated using HeatGen at (a) $\overline{T}_\textrm{fixed}=550  \textrm{K}$, (b) $\overline{T}_\textrm{fixed}=500  \textrm{K}$, (c) $\overline{T}_\textrm{fixed}=475  \textrm{K}$.}
    \label{fig:2DCFD}
\end{figure}

\begin{figure}[H]
    \centering
       \includegraphics[width=0.95\textwidth]{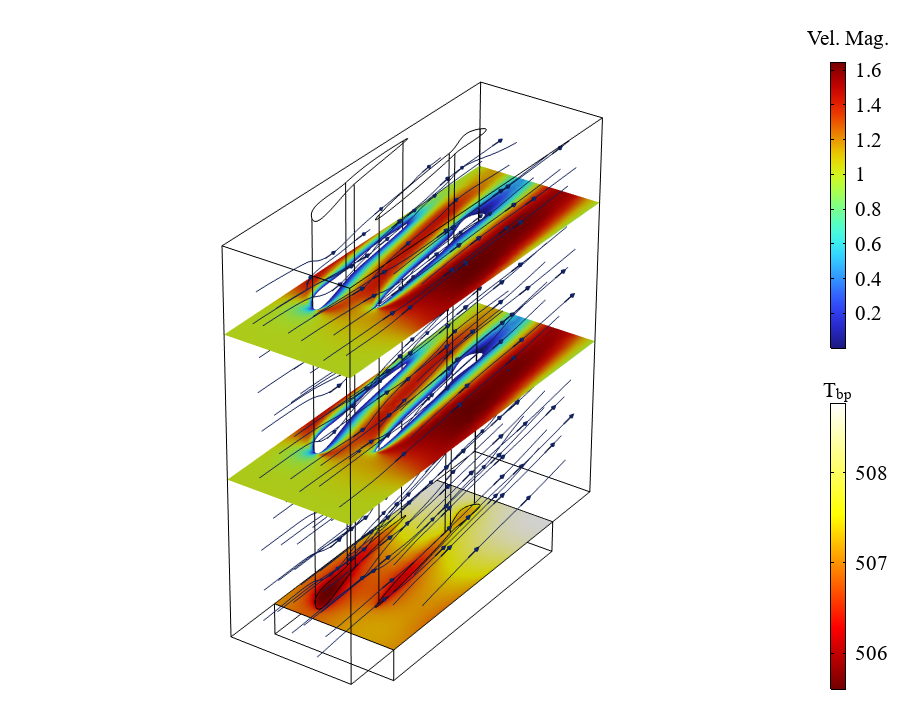}
    \caption{A full 3D visualization of the flow field (illustrated by the streamlines and velocity distribution in the top two cut planes) and the corresponding temperature distribution in the base plate (shown in the bottom cut plane) for the optimized geometry generated by HeatGen at $\overline{T}_\textrm{fixed}=500 \textrm{ K}$.}
    \label{fig:3d500}
\end{figure}

\begin{figure}[H]
    \centering
       \includegraphics[width=0.95\textwidth]{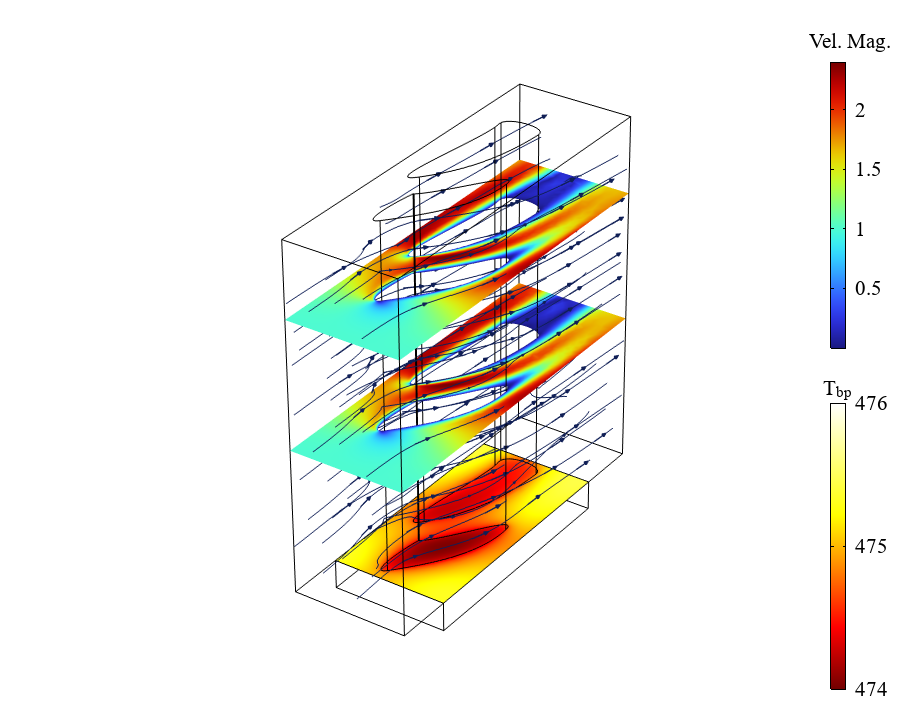}
    \caption{A full 3D visualization of the flow field (illustrated by the streamlines and velocity distribution in the top two cut planes) and the corresponding temperature distribution in the base plate (shown in the bottom cut plane) for the optimized geometry generated by HeatGen at $\overline{T}_\textrm{fixed}=475 \textrm{ K}$.}
    \label{fig:3d475}
\end{figure}

To further validate the optimized designs, full CFD simulations were conducted under identical boundary and operating conditions. The velocity and temperature profiles for the designs generated with maximum temperatures of \(T = 475~\text{K}\), \(500~\text{K}\), and \(550~\text{K}\) are shown in Fig.~\ref{fig:2DCFD}. The CFD results confirm that the designs generated by HeatGen exhibit not only low pressure drops but also maintain surface temperatures within the prescribed limits, demonstrating the effectiveness of the guided diffusion framework in achieving physically consistent and high-performance thermal designs. Full 3D simulation results are also shown in Fig.~\ref{fig:3d500} and Fig.~\ref{fig:3d475} for \(T = 500~\text{K}\) and \(T = 475~\text{K}\), respectively. The velocity and temperature distributions across different fin heights are presented to further illustrate the design performance and the validation of constraints.

\section{Conclusion}
\label{sec:conc}
In this study, we presented a generative design framework for optimizing heat sink geometries under constraint-based objectives aimed at reducing surface temperature and minimizing pressure drop. The approach integrates a denoising diffusion probabilistic model (DDPM) with gradient-based guidance computed using the PyTorch autograd function to generate thermally efficient and physically consistent designs. The DDPM is trained on geometric features derived from boundary representations and guided by the objective values and their gradients with respect to the design vector, computed by surrogate models that are themselves trained on data from pseudo–three-dimensional simulations. The gradients enable the framework to learn an inverse mapping between geometric configurations and their corresponding thermo-fluidic performances. Using this learned mapping, the inference process of the trained DDPM is guided to generate multiple fin geometries whose shapes and coordinates satisfy a specified temperature constraint while minimizing pressure drop as a secondary objective. To assess generalization, inference was performed across a wide range of maximum temperature constraints, from \(400~\mathrm{K}\) to \(550~\mathrm{K}\). Across nearly all temperature limits, the proposed diffusion-guided framework consistently outperformed CMA-ES optimization, achieving up to \(10\%\) lower pressure drops while satisfying maximum temperature constraints. The denoising process from noisy fin shapes to optimized configurations was visualized across diffusion timesteps. The impact of the guidance parameters on the feasibility rate was also evaluated, showing that more than 98\% of the generated designs were feasible. These designs were further validated using full computational fluid dynamics simulations in COMSOL Multiphysics at different constraint temperature values. These results highlight the robustness, scalability, and ability of the framework to efficiently adapt to new thermal conditions.

Overall, this work demonstrates the potential of diffusion-based generative modeling for scalable and data-driven thermal design. Once trained, the DDPM can rapidly generate new design samples at negligible computational cost, making it suitable for integration into digital-twin environments and real-time thermal management systems. Future extensions will explore reinforcement learning–based reward guidance, the inclusion of manufacturing constraints, and multiphysics coupling (e.g., phase change and two-phase cooling) to further advance the development of general-purpose, physics-aware generative design frameworks.

\section*{Acknowledgments}
The authors would like to acknowledge the Natural Sciences and Engineering Research Council of Canada (NSERC) for supporting this research through an NSERC Alliance grant in collaboration with Elomatic Ltd. and InnovMarine Inc. The research was enabled in part by computational resources and services provided by the Digital Research Alliance of Canada and the Advanced Research Computing facility at the University of British Columbia (UBC).

\appendix

\section{Construction of Closed Bézier Geometries}
\label{app:bezier}
To parameterize the fin boundaries, composite Bézier curves are utilized, which provide for smooth geometric variation while maintaining continuity between segments of the curve. A Bézier curve of degree $n$ with the control points $\mathbf{P}_i$ is defined as:
\begin{equation} \mathbf{B}(t) = \sum_{i=0}^{n} \binom{n}{i}(1-t)^{\,n-i}t^{\,i}\mathbf{P}_i, \qquad t\in[0,1]. \end{equation}
In this study, each fin boundary is created using four cubic Bézier segments. A cubic segment can be represented in terms of
\begin{equation} \mathbf{B}(t)= (1-t)^3\mathbf{P}_0 + 3(1-t)^2t\,\mathbf{P}_1 + 3(1-t)t^2\mathbf{P}_2 + t^3\mathbf{P}_3. \end{equation}
Thus, twelve control points are required to define each fin geometry.
Starting from a polygonal outline with primary $m=4$ vertices per fin, we want to obtain a closed smooth boundary. The primary vertices are generated at perturbed angular and radial positions:
\begin{equation} \theta_i = i\frac{2\pi}{m} + \delta\theta_i\frac{\pi}{m}, \qquad r_i = |\delta r_i|\,r_{\mathrm{max}}, \end{equation} converted to Cartesian coordinates: \begin{equation} x_i = r_i \cos(\theta_i), \qquad y_i = r_i \sin(\theta_i). \end{equation}
The polygon is then translated by $(x_{\mathrm{shift}}, y_{\mathrm{shift}})$ 
to locate the fin within the design space.
Each polygon edge is replaced by a cubic Bézier curve. To obtain smooth transitions, a tangent direction is assigned to each main vertex $\mathbf{P}_i$ by taking a weighted average of the outgoing and incoming edge directions:
\begin{equation} \tilde{\phi}_i = w_i\phi_i^{\mathrm{out}} + (1-w_i)\phi_{i-1}^{\mathrm{out}}, \qquad w_i = 0.5(1+\eta_i), \quad \eta_i\in[0,1]. \end{equation}
For an edge $\mathbf{P}_0 \rightarrow \mathbf{P}_3$ with chord length $d = \|\mathbf{P}_3 - \mathbf{P}_0\|$, the interior control points are positioned along the tangent directions at a distance:
\begin{equation} r_m = 0.707\, r_{\mathrm{mid}}\, d, \qquad r_{\mathrm{mid}}\in[0,1], \end{equation}
resulting in: \begin{equation} \mathbf{P}_1 = \mathbf{P}_0 + r_m(\cos\tilde{\phi}_0, \sin\tilde{\phi}_0), \qquad \mathbf{P}_2 = \mathbf{P}_3 - r_m(\cos\tilde{\phi}_3, \sin\tilde{\phi}_3). \end{equation}
The higher values of $r_{\mathrm{mid}}$ produce stronger curvature, while the smaller values produce flatter segments. This construction can be repeated around the polygon in order to create a smooth, closed composite Bézier contour for each fin.
\subsection{Parameter Vector for Shape Generation and Variation}\label{sec:parameter-vector}
The geometry of the fins is controlled by a parameter vector $X$ , which encodes both the placement and shape of each fin. For $n_f$ fins, each defined by $m$ primary outline vertices, the total number of design parameters is:
\begin{equation}
\dim(X) = 2n_f + 3mn_f
\end{equation}
The first $2n_f$ components define the translational changes of each fin centroid in the $x$ and $y$ directions:
\begin{equation}
X[0:n_f) \rightarrow x\text{-shifts}, \qquad
X[n_f:2n_f) \rightarrow y\text{-shifts}.
\end{equation}
The remaining $3mn_f$ components define the fin boundary shape. For every one of the $m$ primary vertices of each fin, the vector includes: $\delta r$: radial deformation (controls vertex distance from the centroid), $\delta\theta$: angular perturbation (rotational displacement), $\eta$: curvature weighting (controls smoothness of the Bézier segment transitions).
Thus, the parameter vector has the structured form:
\begin{equation} X = \underbrace{[x_{1}, \dots, x_{n_f}, \; y_{1}, \dots, y_{n_f}]}_{\text{Fin placement}} \;\cup\; \underbrace{[\delta r, \delta \theta, \eta]_{1,1}, \dots, [\delta r, \delta \theta, \eta]_{m, n_f}}_{\text{Fin shape variation}}. \label{eq:param_vector} \end{equation}
These parameters are then utilized to construct the fin geometry based on the composite Bézier-based procedure discussed in the previous section.

\section{DDPM Parameters and Hyperparameters }\label{app:ddpm}

\begin{table}[H]
\centering
\caption{Hyperparameters and training configuration of the 1D U-Net DDPM used for generative modeling of heat sink design vector}
\begin{tabular}{l l}
\hline
\textbf{Component} & \textbf{Value} \\
\hline
Diffusion steps $T$ & $1000$ \\
Noise schedule & cosine \\
Base channels & $64$ \\
Time embedding & $256$ (sinusoidal) \\
Dropout & $0.1$ \\
Batch size & $128$ \\
Epochs & $750$ \\
Optimizer & AdamW \\
Learning rate & $1\times10^{-4}$ \\
Weight decay & $0$ \\
Loss & MSE \\
\hline
\end{tabular}
\end{table}

\section{Regression models architecture}\label{app:modelap}

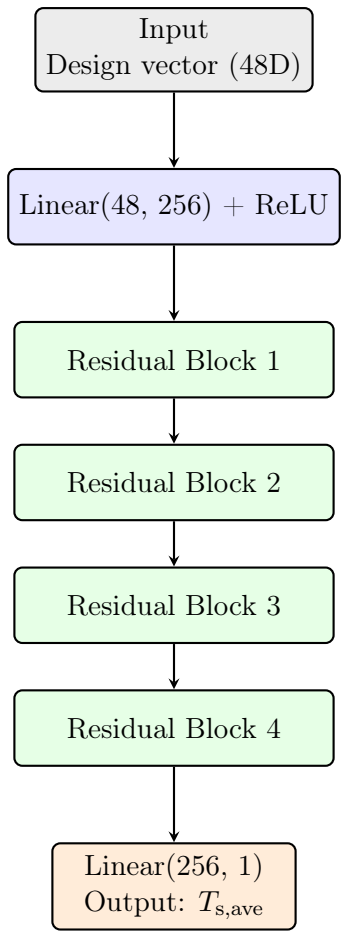
\begin{figure}[H]
\centering
\begin{tikzpicture}[>=stealth, thick, font=\small, node distance=1.0cm and 1.3cm]

\node[draw, rounded corners=3pt, fill=gray!15,
      minimum width=3.2cm, minimum height=1cm]
  (input) {\shortstack{Input\\Design vector (48D)}};

\node[draw, rounded corners=3pt, fill=blue!10, below=of input, minimum width=3.4cm, minimum height=1cm]
  (fc1) {Linear(48, 256) + ReLU};

\node[draw, rounded corners=3pt, fill=green!10, below=1.0cm of fc1, minimum width=4.2cm, minimum height=1cm]
  (rb1) {Residual Block 1};
\node[draw, rounded corners=3pt, fill=green!10, below=0.6cm of rb1, minimum width=4.2cm, minimum height=1cm]
  (rb2) {Residual Block 2};
\node[draw, rounded corners=3pt, fill=green!10, below=0.6cm of rb2, minimum width=4.2cm, minimum height=1cm]
  (rb3) {Residual Block 3};
\node[draw, rounded corners=3pt, fill=green!10, below=0.6cm of rb3, minimum width=4.2cm, minimum height=1cm]
  (rb4) {Residual Block 4};

\node[draw, rounded corners=3pt, fill=orange!15,
      below=1.0cm of rb4,
      minimum width=3.2cm, minimum height=1cm]
  (out) {\shortstack{Linear(256, 1)\\Output: $T_{\mathrm{s,ave}}$}};

\draw[->] (input) -- (fc1);
\draw[->] (fc1) -- (rb1);
\draw[->] (rb1) -- (rb2);
\draw[->] (rb2) -- (rb3);
\draw[->] (rb3) -- (rb4);
\draw[->] (rb4) -- (out);

\end{tikzpicture}
\caption{%
ResNet architecture for predicting the average surface temperature from 48 geometry parameters.
}
\label{fig:resmlp_layers}
\end{figure}

\begin{table}[h!]
\centering
\caption{Temperature regression model parameters}
\begin{tabular}{l l}
\hline
\textbf{Parameter} & \textbf{Specification} \\
\hline
Input dimension & 48D (geometry vector) \\
Hidden dimension & 256 neurons \\
Residual block & Linear(256,256) + ReLU + BatchNorm + skip connection \\
Optimizer & AdamW \\
Learning rate & $5\times10^{-4}$ \\
Weight decay & $1\times10^{-4}$ \\
Batch size & 256 \\
Loss function & Mean squared error (MSE)  \\
\hline
\end{tabular}
\label{tab:temperature_resmlp_config}
\end{table}

\begin{figure}[H]
\centering
\begin{tikzpicture}[>=stealth, thick, font=\small, node distance=1.0cm and 1.3cm]

\node[draw, rounded corners=3pt, fill=gray!15,
      minimum width=3.2cm, minimum height=1cm]
  (input) {\shortstack{Input\\Design vector (48D)}};

\node[draw, rounded corners=3pt, fill=blue!10, below=of input, minimum width=3.6cm, minimum height=1cm]
  (fc1) {Linear(48, 512) + ReLU};

\node[draw, rounded corners=3pt, fill=green!10, below=1.0cm of fc1, minimum width=4.5cm, minimum height=1cm]
  (rb1) {Residual Block 1};
\node[draw, rounded corners=3pt, fill=green!10, below=0.6cm of rb1, minimum width=4.5cm, minimum height=1cm]
  (rb2) {Residual Block 2};
\node[draw, rounded corners=3pt, fill=green!10, below=0.6cm of rb2, minimum width=4.5cm, minimum height=1cm]
  (rb3) {Residual Block 3};
\node[draw, rounded corners=3pt, fill=green!10, below=0.6cm of rb3, minimum width=4.5cm, minimum height=1cm]
  (rb4) {Residual Block 4};

\node[draw, rounded corners=3pt, fill=orange!15,
      below=1.0cm of rb4,
      minimum width=3.2cm, minimum height=1cm]
  (out) {\shortstack{Linear(512, 1)\\Output: $p_{\mathrm{pred}}$}};

\draw[->] (input) -- (fc1);
\draw[->] (fc1) -- (rb1);
\draw[->] (rb1) -- (rb2);
\draw[->] (rb2) -- (rb3);
\draw[->] (rb3) -- (rb4);
\draw[->] (rb4) -- (out);
\end{tikzpicture}

\caption{%
ResNet architecture for predicting pressure drop from 48 geometry parameters.
}
\label{fig:pressure_resmlp_layers}
\end{figure}
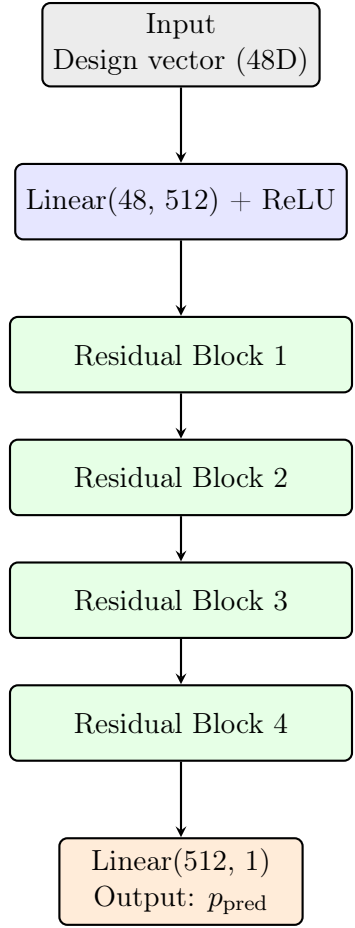

\begin{table}[H]
\centering
\caption{Pressure regression model parameters}
\begin{tabular}{l l}
\hline
\textbf{Parameter} & \textbf{Specification} \\
\hline
Input dimension & 48D (geometry vector) \\
Hidden dimension & 512 neurons \\
Dropout & 0.1\\
Residual block & Linear(512,512) + ReLU + BatchNorm + skip connection \\
Optimizer & AdamW \\
Learning rate & $1\times10^{-4}$ \\
Weight decay & $1\times10^{-4}$ \\
Batch size & 512 \\
Loss function & Mean squared error (MSE)  \\
\hline
\end{tabular}
\label{tab:pressure_resmlp_config}
\end{table}

\bibliographystyle{elsarticle-num}
\bibliography{cas-refs}







\end{document}